\documentclass[10pt,twocolumn,letterpaper]{article}

\usepackage[pagenumbers]{cvpr} 
\usepackage{amssymb}

\definecolor{cvprblue}{rgb}{0.21,0.49,0.74}
\usepackage[pagebackref,breaklinks,colorlinks,allcolors=cvprblue]{hyperref}

\def\paperID{14367} 
\def\confName{CVPR}
\def\confYear{2025}

\title{
A Noise is Worth Diffusion Guidance
}

\author{
Donghoon Ahn$^{\bullet1}$ \hspace{0.03\linewidth} 
\and
Jiwon Kang$^{\bullet1}$ \hspace{0.03\linewidth}
\and
Sanghyun Lee$^{\circ2}$
\and
Jaewon Min$^{\circ1}$ \hspace{0.04\linewidth}
\and
Minjae Kim$^1$\vphantom{$^{\dagger}$}
\and
Wooseok Jang$^1$
\and
Hyoungwon Cho$^1$
\and
Sayak Paul$^4$
\and
SeonHwa Kim$^1$\hspace{0.04\linewidth}
\and
Eunju Cha\textsuperscript{$\dagger3$}
\and
Kyong Hwan Jin\textsuperscript{$\dagger1$}
\and
Seungryong Kim\textsuperscript{$\dagger2$}
}

\usepackage{lipsum}
\usepackage{hyperref}
\usepackage{url}
\usepackage{booktabs}
\usepackage{multirow}
\usepackage{wrapfig}
\usepackage{xcolor,colortbl}
\usepackage{algpseudocode}
\usepackage{amsmath}
\usepackage{colortbl}
\usepackage{xcolor}
\usepackage{amssymb}
\usepackage[utf8]{inputenc}
\usepackage{float}
\usepackage{svg}
\usepackage{pifont}
\usepackage{graphicx}

\usepackage{amsthm}
\usepackage{xspace}

\newcommand{\ours}{\textit{\textbf{NoiseRefine}}\xspace}
\newcommand{\ourmodel}{noise refining model\xspace}
\newcommand{\cfgimage}{x_{0}^{\text{Guide}}}
\newcommand{\invnoise}{x_{T}^{\text{Guide}}}
\newcommand{\trueinvnoise}{x^{\text{Guide}\dagger}_{T}}
\newcommand{\predimage}{\hat{x}_0}
\newcommand{\prednoise}{\hat{x}_T}

\newcommand{\citeguidance}{\citep{ho2022classifier,ahn2024self,hong2023improving,sadat2024no,hong2024smoothed,karras2024guiding}\xspace}
\newcommand{\citeinversion}{\citep{song2020denoising,meiri2023fixed,garibi2024renoise}\xspace}
\newcommand{\paragrapht}[1]{\vspace{-10pt}\paragraph{#1}}

\begin{document}
\twocolumn[{%
\renewcommand\twocolumn[1][]{#1}%
\maketitle
\begin{center}
    \vskip -2em
    {\normalsize
        $^1$Korea University \hspace{0.05\linewidth}
        $^2$KAIST \hspace{0.05\linewidth}
        $^3$Sookmyung Women's University \hspace{0.05\linewidth}
        $^4$Hugging Face
    \par}
    \vskip 1em
\end{center}
\begin{center}
    \centering
    \captionsetup{type=figure}
    \includegraphics[width=1\textwidth]{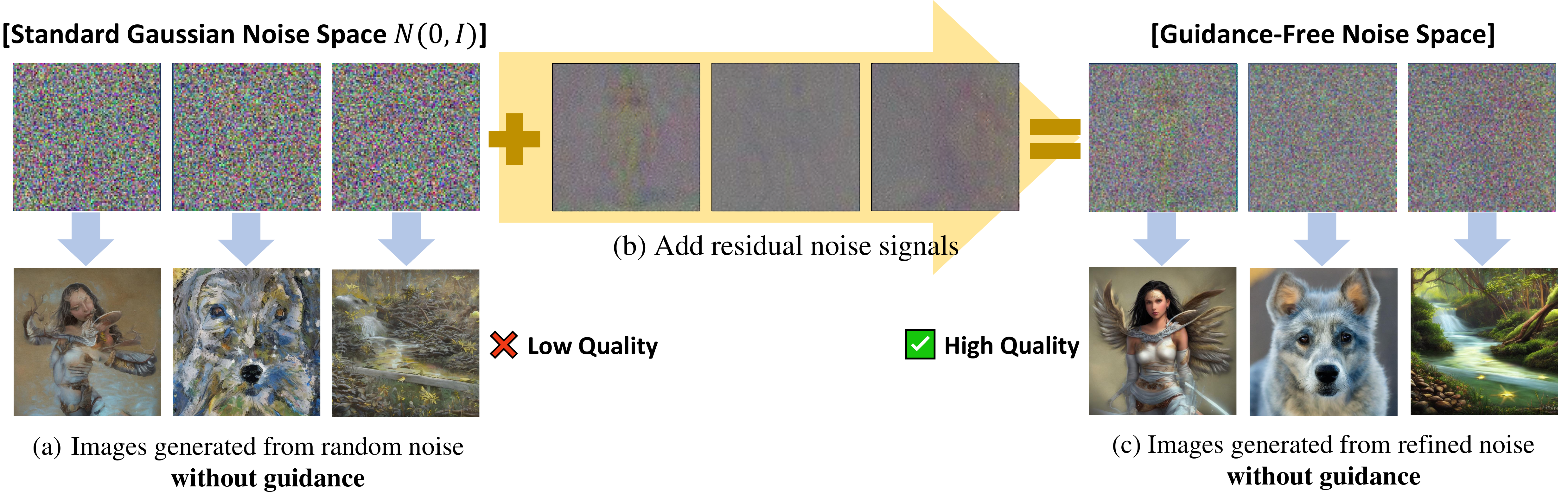}
    \vspace{-15pt}
    \captionof{figure}{\textbf{Effectiveness of \ours.} Diffusion models often fail to generate high-quality images without guidance, such as classifier-free guidance (CFG)~\cite{ho2022classifier}. We propose \textbf{\ours}, a novel approach to improve image quality without use of guidance by learning to map initial random noise to a guidance-free noise space. Results are demonstrated using Stable Diffusion 2.1\citep{rombach2022high}.}
    \label{fig:teaser}
\vspace{2mm}
\end{center}%
}]

\begingroup
\renewcommand{\thefootnote}{}
\footnotetext{\scriptsize $\bullet$,$\circ$: Equal contribution, \scriptsize $\dagger$: Co-corresponding author}
\endgroup

\begin{abstract}
Diffusion models excel in generating high-quality images. However, current diffusion models struggle to produce reliable images without guidance methods, such as classifier-free guidance (CFG). Are guidance methods truly necessary? Observing that noise obtained via diffusion inversion can reconstruct high-quality images without guidance, we focus on the initial noise of the denoising pipeline. By mapping Gaussian noise to `guidance-free noise', we uncover that small low-frequency components significantly enhance the denoising process, removing the need for guidance and thus improving both inference throughput and memory. Expanding on this, we propose \ours, a novel method that replaces guidance methods with a single refinement of the initial noise. This refined noise enables high-quality image generation without guidance, within the same diffusion pipeline. Our noise-refining model leverages efficient noise-space learning, achieving rapid convergence and strong performance with just 50K text-image pairs. We validate its effectiveness across diverse metrics and analyze how refined noise can eliminate the need for guidance. See our project page: 
\href{https://cvlab-kaist.github.io/NoiseRefine/}{https://cvlab-kaist.github.io/NoiseRefine/}.
\vspace{-5pt}

\end{abstract}

\section{Introduction}
\label{sec:intro}

In recent years, Text-to-Image (T2I) diffusion models~\citep{rombach2022high,esser2024scaling,chen2023pixart,podell2023sdxl,saharia2022photorealistic}, which generate images conditioned on text prompts, have achieved remarkable advancements. However, their ability to produce high-quality samples largely relies on guidance techniques, such as classifier-free guidance (CFG)~\citep{ho2022classifier} and its variants~\citep{dhariwal2021diffusion,ahn2024self,hong2023improving,hong2024smoothed}. These methods significantly enhance image quality during inference but double the computational cost. Despite drawbacks such as increased batch size, high guidance scale requirements, oversaturation, and reduced diversity, the dramatic performance gains make CFG the de facto standard. Recent works~\citep{sadat2023cads,chung2024cfg++,sadat2024no} aim to mitigate these limitations, but the impact of CFG on image quality makes it indispensable in most diffusion pipelines.

This raises a fundamental question: \textit{Can we replace the effects of guidance techniques with minimal changes to the diffusion pipeline?} While some works have proposed distilling classifier-free guided scores into student models~\citep{meng2023distillation,lin2024sdxl,kang2025distilling}, these methods require extensive training data, significant computational resources, and large model capacities to replicate guided denoising trajectories. Instead, we explore the inputs of T2I diffusion pipelines: a prompt and an initial noise. A single prompt can yield diverse samples depending on the initial noise. In addition, recent studies~\citep{samuel2024generating, xu2024good} highlighted the strong influence of initial noise on output quality, a correlation between the starting noise and the resulting image. 
In fact, we observe that, on rare occasions, certain random initial noises can produce high-quality images.
This implies that if we could find such noise easily, we would successfully eliminate the need for guidance methods. Thus, we aim to find a noise space capable of generating high-quality images, which we term the \textit{`guidance-free noise space'}.

Throughout this work, we explore how we can find this guidance-free noise space. Some works have proposed to select or optimize noise space to improve perceptual quality~\citep{eyring2024reno,qi2024not} or prompt adherence~\citep{guo2024initno}. However, those are not intended to replace the guidance techniques and then still rely on them. Furthermore, they typically require extensive iterations to optimize the input noise~\citep{eyring2024reno,qi2024not,guo2024initno} and often only work in the few-step diffusion model~\citep{eyring2024reno}.

\begin{figure}
    \centering
    \includegraphics[width=\linewidth]{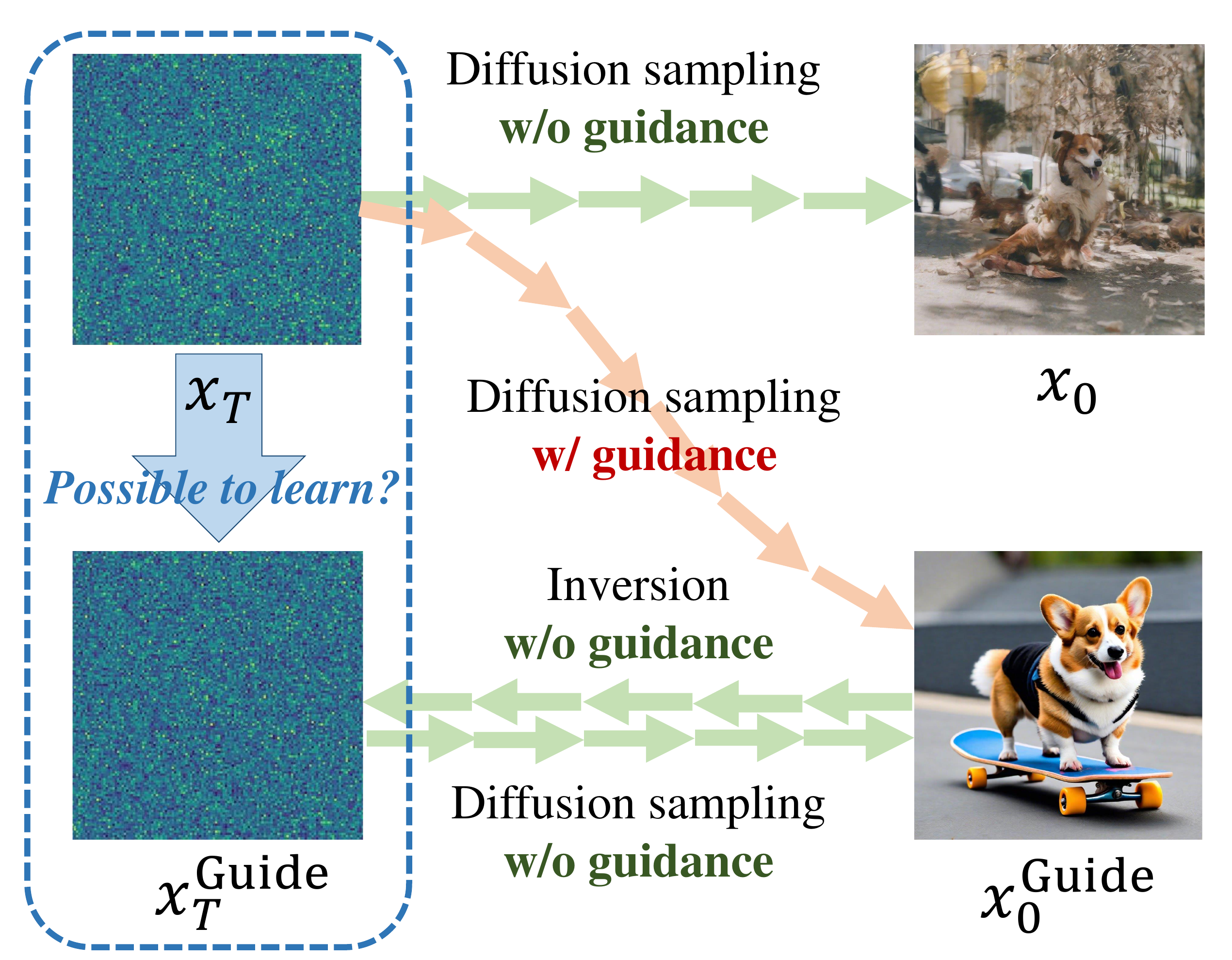}
    \caption{\textbf{Insight of \textbf{\ours}.} We combine inversion methods~\citeinversion and guidance methods~\citeguidance to establish a mapping between standard noise  \( x_T \) and guidance-free noise $\invnoise$.}
    \label{fig:idea}
\vspace{-10pt}
\end{figure}

Our key insight to find a guidance-free noise space lies in leveraging diffusion inversion methods~\citep{song2020denoising,garibi2024renoise,meiri2023fixed} to obtain initial noises that are reconstructed to their corresponding high-quality images without guidance. Building on this, we generate multiple high-quality images using guidance techniques~\citep{ho2022classifier,ahn2024self} and apply inversion to these images. This process yields a collection of initial noises capable of reconstructing images without guidance, constructing the guidance-free noise space. The overall concept is illustrated in Fig.~\ref{fig:idea}. 

We analyze the difference between inversion noise and standard Gaussian noise, and we find that the difference arises primarily from subtle variations in pixel values and is concentrated in the low-magnitude, low-frequency component. Our objective is for the noise refining model to learn the mapping from standard Gaussian noise to `guidance-free noise'.



Although we could directly learn the mapping between the initial noises and the inversion noises, error accumulation during the inversion process~\cite{garibi2024renoise,meiri2023fixed} makes this approach suboptimal.
Thus, we mitigate this inversion error by shifting the distance loss from noise space to its denoised image space. We refer to our method as \textbf{\ours}. Fig.~\ref{fig:method} illustrates the overall training process of \ours. 


Training \ourmodel by backpropagating gradients through the full denoising steps can result in substantial memory consumption and unstable training due to multi-step gradient propagation. To address this challenge, we propose multistep score distillation (MSD), a simple yet effective technique enabling efficient full-step model optimization without incurring the high cost of backpropagation. Inspired by score distillation sampling (SDS)~\citep{poole2022dreamfusion}, MSD skips gradient computation within the denoising network during the denoising process. Notably, we found that skipping a computation of gradient not only reduces computational overhead but also accelerates model convergence.

With just prompts and the basic diffusion model, our approach easily accomplish self-distillation without the need for natural images. We demonstrate that sampling from the noise refined by our model produces high-quality images without guidance, as validated across various benchmarks. These results are comparable to images generated using CFG~\citep{ho2022classifier} and PAG~\citep{ahn2024self} on the same diffusion model while being approximately twice as fast (compared to CFG alone) or three times as fast (compared to both CFG and PAG).

Our contributions are summarized as:
\begin{itemize}
\item We identify the existence of a noise space that enables high-quality generation \textit{without} guidance~\citep{ho2022classifier,ahn2024self}, which we refer to as `guidance-free noise space'.


\item We show that the mapping between a standard normal distribution and guidance-free noise space can be efficiently learned by a neural network.

\item To reduce backpropagation costs in training, we propose Multistep Score Distillation that detaches gradients during denoising, accelerating convergence.

\item Our approach achieves a 2x speed-up compared to using guidance methods, maintaining comparable quality.
\end{itemize}

\begin{figure}
    \centering
    \includegraphics[width=\linewidth]{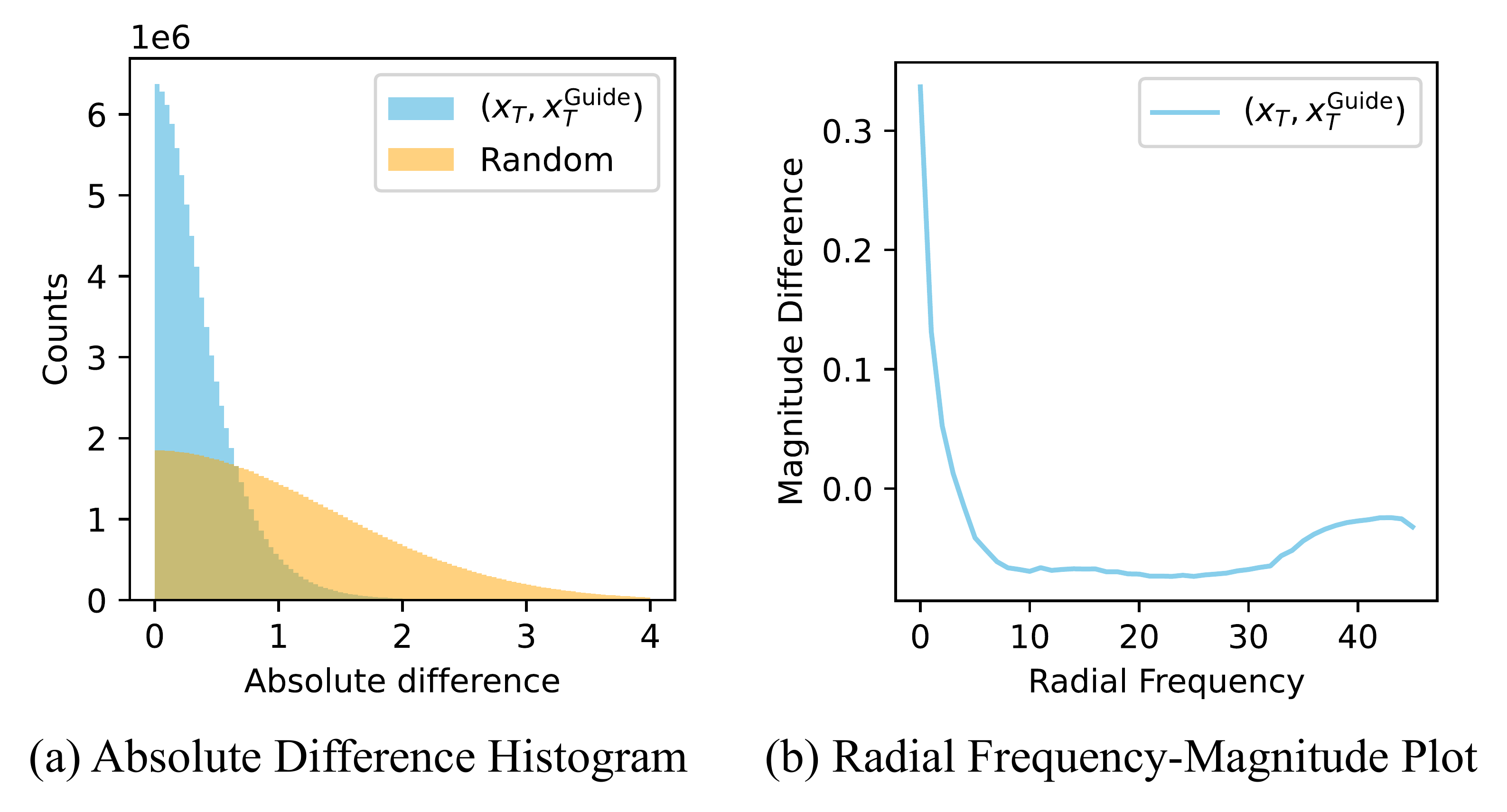}
    \vspace{-20pt}
    \caption{
    \textbf{Analysis on the relationship between the initial Gaussian noise} \( x_T \) \textbf{and the guidance-free noise} \( \invnoise \) \textbf{.} (a) shows the histogram of the absolute difference between \( x_T \) and \( \invnoise \). Here, `Random' denotes the setting where the both noises are replaced with independent gaussian white noise.  (b) presents the magnitude difference between the 2D Fourier-transformed frequency components of \( \mathcal{F}(x_T) \) and \( \mathcal{F}(\invnoise) \). The difference between \( x_T \) and \( \invnoise \) is significantly smaller than in the random case, which mainly corresponds to the low-frequency components.
    }
    \label{fig:analysis}

\vspace{-10pt}
\end{figure}
\vspace{-5pt}
\section{Related Work}

\paragraph{Diffusion guidance.}
Classifier Guidance (CG)~\citep{mao2023guided} enhances fidelity by leveraging trained classifier gradients, albeit at the cost of diversity.    CFG~\citep{ho2022classifier} models an implicit classifier to achieve similar effects. Ahn et al.~\citep{ahn2024self} and Karras et al.~\citep{karras2024guiding} further generalize those guidance methods by intentionally generating lower-quality samples to guide the process toward improved outputs and other guidance techniques~\citep{hong2023improving,sadat2024no,hong2024smoothed} generate `bad' samples in various ways. While effective, these methods double computational and memory costs by requiring degraded sample generation at each step, which is essential to their operation.

\paragrapht{Diffusion inversion.}
Denoising Diffusion Implicit Models (DDIM)~\citep{song2020denoising} introduced deterministic sampling, enabling inversion from image to noise. This means that by starting the sampling process from the inverted noise, we can reconstruct the original images. Although DDIM Inversion~\citep{song2020denoising} is the most commonly used inversion method for diffusion models, its reliance on linear approximation often leads to noticeable artifacts in reconstructed images. Several works~\citep{pan2023effective,garibi2024renoise,meiri2023fixed} employ fixed-point iteration to reduce the approximation error.  If guidance is used during inversion, then guidance must be applied during sampling to achieve the same generated image and the same holds in reverse.
\vspace{-5pt}

\paragrapht{Noise optimization.}
Optimizing or selecting noises with certain objectives has been a key research focus in diffusion models~\citep{samuel2024generating, eyring2024reno, mao2023semantic}. ReNO~\citep{eyring2024reno} optimizes noises based on reward models and Samuel et al.~\citep{samuel2024generating} proposed a bootstrap-based method to optimize initial noises for rare concept generation. However, these optimization methods require a substantial number of iterations, which poses a challenge for real-world applications.  Another approach~\citep{mao2023semantic} involves constructing a noise database to generate initial noise during inference but is not generalized to unseen prompts. Our approach overcomes both limitations by learning a direct mapping to the learned noise space, enabling efficient mapping with a single inference step and providing generalization power for new prompts.


\section{Method}
In this section, we first \textbf{identify characteristics of mapping from the Gaussian noise space to guidance-free noise space}, a space of initial noises that can be denoised into high-quality images without guidance (Sec.~\ref{subsec:guidance-free_noise_space}). Next, we \textbf{introduce a method for learning this mapping from arbitrary Gaussian noise} (Sec.~\ref{subsec:noisegen}). Finally, we demonstrate that, with a carefully designed dataset construction and filtering process, predicting guidance-free noise using a simple single-step neural network can effectively replace traditional guidance techniques~\citep{ho2022classifier,ahn2024self}, enabling efficient and high-quality image generation without guidance (Sec.~\ref{subsec:asthetic_filtering}).

\subsection{Guidance-Free Noise Space}
\label{subsec:guidance-free_noise_space}

To obtain the guidance-free noise space, we emphasize the capability of inversion methods~\citep{song2020denoising,garibi2024renoise,meiri2023fixed} to precisely reconstruct the original image without guidance. In theory, inverting an infinite set of natural images would fully capture this space, but this is infeasible. Instead, we leverage the powerful generation capabilities of text-to-image diffusion models~\citep{rombach2022high}, which produce high-quality images with guidance~\citep{ho2022classifier,ahn2024self}.

Specifically, Gaussian noise $x_T$ is sampled from standard Gaussian distribution $\mathcal{N}(0, I)$ and denoised into a plausible image \( \cfgimage \), using text prompt or condition $c$ with CFG~\citep{ho2022classifier} or any other guidance method~\citep{ahn2024self,hong2024smoothed}. Inverting the image with an inversion method~\citep{garibi2024renoise,meiri2023fixed} gives us the noise \( \invnoise \), defined as:
\begin{equation}
\invnoise := \text{Inversion}(\text{Denoise}^{\text{Guide}}(x_T,c)),
\end{equation}
where $\text{Inversion}(\cdot)$ and $\text{Denoise}^{\text{Guide}}(\cdot)$ denote inversion~\citep{garibi2024renoise,meiri2023fixed} and denoising with condition $c$ and guidance, respectively. A more detailed explanation of the notations can be found in supplementary material~\ref{supp:preliminaries}. Note that the generated image $\cfgimage$ and inversion noise $\invnoise$ are conditioned on context $c$ such as the text prompt, but we omit the notation $c$ in the paper for simplicity. Now, we can map $x_T$ into $\cfgimage$. Ideally, if the mapping is consistent or generalizable, a neural network can learn to map initial noise to guidance-free noise. This concept is illustrated in Fig.~\ref{fig:idea}. 

We investigate the structure of this mapping by generating \( \{x_T, \invnoise\} \) pairs via the aforementioned process with 10K randomly selected prompts from the MS-COCO dataset~\citep{lin2014microsoft}. We employ Stable Diffusion 1.5~\citep{rombach2022high}, CFG~\citep{ho2022classifier}, and ReNoise~\citep{garibi2024renoise}.
Comparing the absolute differences between \( x_T \) and \( \invnoise \) to those between random noise instances, Fig.~\ref{fig:analysis} (a) shows that the differences in \( \{x_T, \invnoise\} \) pairs are significantly smaller than those of `Random' pairs. These differences mainly correspond to low-frequency components in the frequency domain as shown in Fig.~\ref{fig:analysis}, which plots the magnitude differences between Fourier-transformed noises. This suggests that guiding initial noise with suitable small low-frequency components for given condition $c$ can generate high-quality samples without additional guidance during the sampling stage.

\begin{figure}[t]
    \centering
    \includegraphics[width=\linewidth]{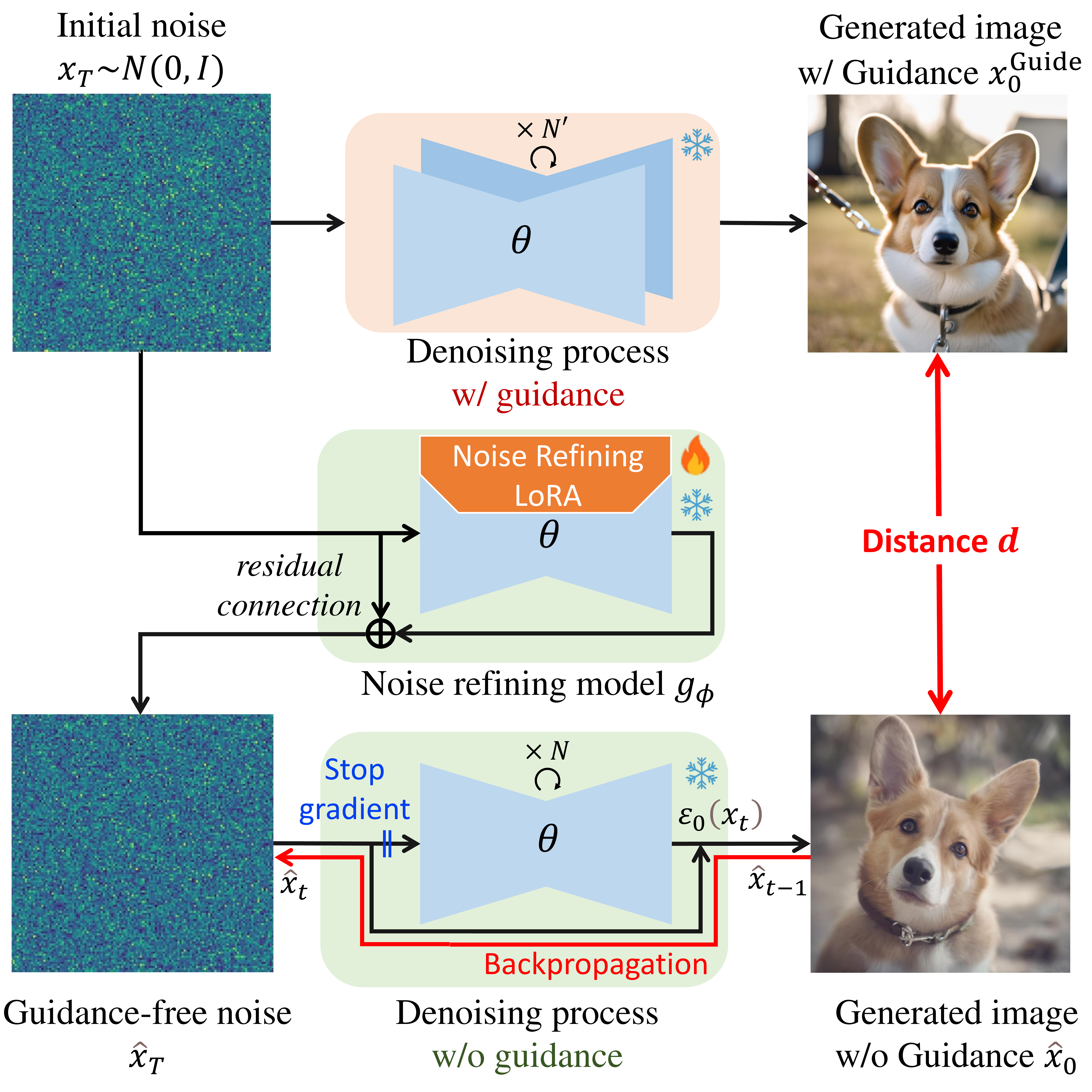}
    \vspace{-15pt}
    \caption{\textbf{Training pipeline.}
    We propose a training methodology to learn a mapping from initial noise to guidance-free noise. Given an initial Gaussian noise \( x_T \), the original diffusion model parameterized by $\theta$ generates an image \( \cfgimage \) using guidance~\citep{ho2022classifier,ahn2024self}. Noise refining model refines the initial noise \( x_T \) to produce $\prednoise = g_\phi(x_T)$, which is then input to the original model to generate an image $\predimage$ without guidance. By minimizing the distance between two images $d(\cfgimage, \predimage)$, \ourmodel effectively learns the desired mapping. Note that both \ourmodel and original model also receive a prompt \( c \) as input, though this is omitted here for simplicity.
    }
    \label{fig:method}

\vspace{-10pt}
\end{figure}


\subsection{Learning to Map Guidance-Free Noise Space}
\label{subsec:noisegen}
\paragraph{Mitigating inversion error.}
\label{subsec:mitigating}
A straightforward approach for learning a mapping to guidance-free noise space would be to learn the inversion noise directly. Although possible enough, inversion methods~\citeinversion have inherent limitations. They rely on approximations, which means \textit{true} inversion noise $\trueinvnoise$ is not guaranteed. Thus, attempting to learn this approximated inversion noise which includes inversion error may limit the performance. Hence, we try to sidestep this issue, by learning directly in the image space. 
To demonstrate the validity of our approach, we investigate whether reducing the image space distance $d(x_0,\cfgimage$) also effectively decreases the noise space distance $d(x_T,\invnoise)$ as training progresses, where $d$ is distance metric measuring the difference between two data points. This relationship is clarified in Proposition~\ref{proposition1} and illustrated in Fig.~\ref{fig:method}. A detailed proof is provided in supplementary material~\ref{supp:derivations}. 

\begin{figure}
    \centering
    \includegraphics[width=1\linewidth]{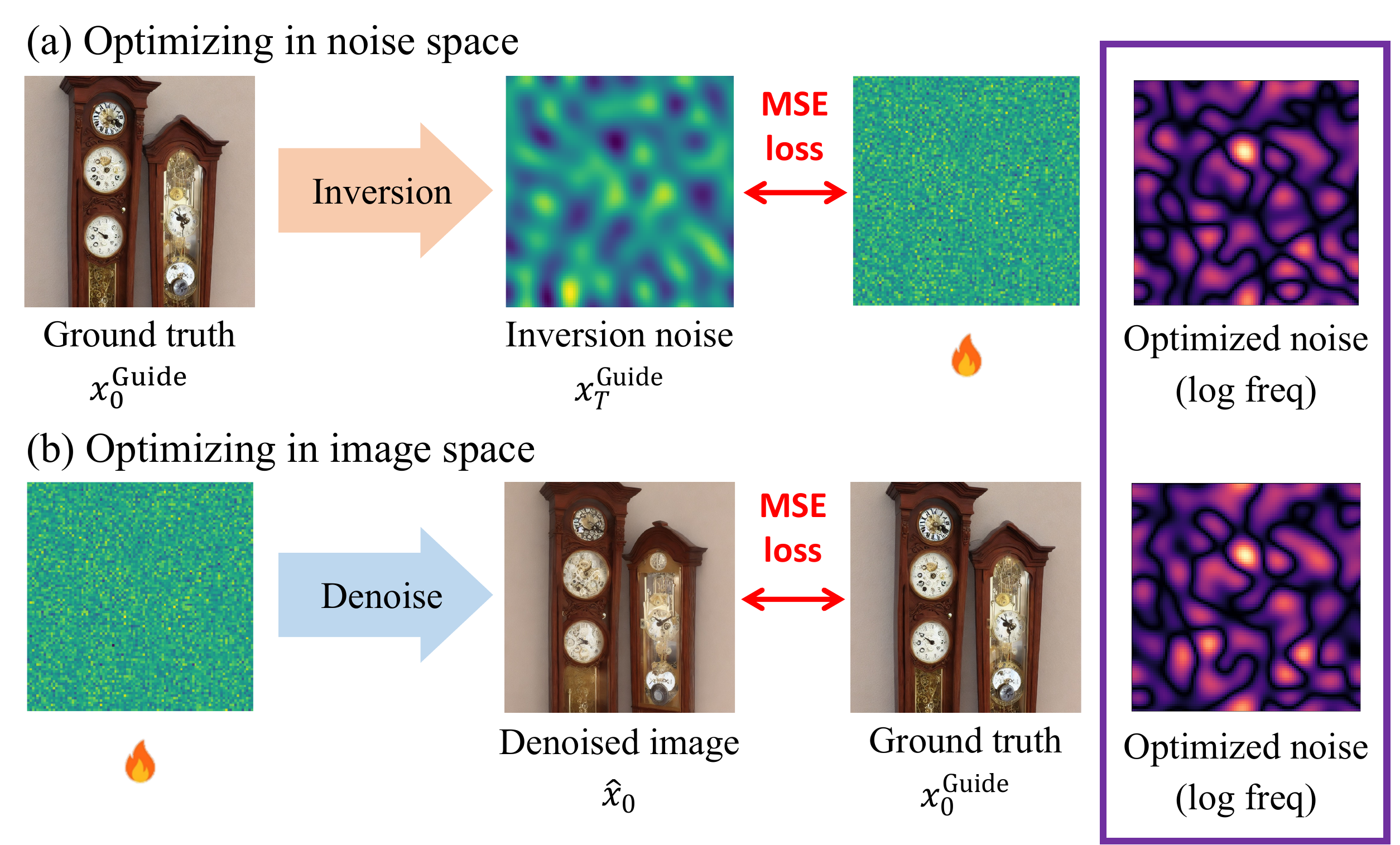}
    \vspace{-20pt}
    \caption{\textbf{Comparison between noise optimization methods.} We compare two methods to optimize a noise for target image generation. (a) illustrates direct optimization using inversion noise from the target image, while (b) shows optimization by minimizing the loss between denoised image and the target image. The rightmost column visualizes each optimized noise in a low-frequency area, indicating the similarity between the two noises.}
    \label{fig:noise_opt_comp}
\vspace{-10pt}
\end{figure}

\newtheorem{proposition}{Proposition}
\begin{proposition}
Let $x_T$ be an initial noise, and suppose that $x_0$ is the image obtained through denoising. Assuming Lipschitz continuity with distance metric $d$, for every $x_T$, there exists a constant $\kappa>0$ such that the following holds:
\[d(x_T,\trueinvnoise)<\kappa d(x_0,\cfgimage).\]
\label{proposition1}
\end{proposition}
\vspace{-15pt}
We support our proposition by conducting toy experiments. In Fig.~\ref{fig:noise_opt_comp}, we compare two strategies: directly optimizing the noise $x_T$ with the empirical inversion noise $\invnoise$, which we treat as true, and optimizing the loss between denoised image $\cfgimage$ and the target image $x_0$. To visualize, we plot the low-frequency regions of two optimized noises, revealing their similarity and supporting our proposition.
\vspace{-10pt}

\paragraph{Training pipeline.} Our overall training framework is illustrated in Fig.~\ref{fig:method}. Given randomly sampled Gaussian noise \( x_T \) and a prompt \( c \), a diffusion pipeline takes the noise $x_T$ as an initial input and generates a guided image \( \cfgimage \) by executing \( N' \) denoising steps with guidance~\citep{ho2022classifier,ahn2024self}. Our \ourmodel $g_\phi(\cdot)$ estimates the refined noise \( \prednoise \). Using this \( \prednoise \) as initial input, the same diffusion pipeline generates an image \( \predimage \) through \( N \) denoising steps \textit{without guidance}. To reduce the gap between \( \predimage \) and \( \cfgimage \), we apply a distance loss $ d(\hat{x}_0, \cfgimage)=\|\hat{x}_0-\cfgimage\|^2_2$. In this way, our model learns to guide the initial noise \( x_T \) toward a guidance-free noise space.


For the architecture of the \ourmodel $g_\phi(\cdot)$, we found that by attaching a lightweight LoRA~\citep{hu2021lora} to the pre-trained diffusion models, the \ourmodel can effectively leverage the diffusion model’s rich knowledge of text and image information, allowing for faster convergence. Additionally, as shown in Fig.~\ref{fig:analysis} (a), the difference between $x_T$ and $\invnoise$ is slight; therefore, we incorporate residual connection~\citep{he2016deep} in the \ourmodel $g_\phi(\cdot)$ to enable the model to converge rapidly during training.

\vspace{-10pt}

\begin{figure}[!t]
    \centering
    \includegraphics[width=\linewidth]{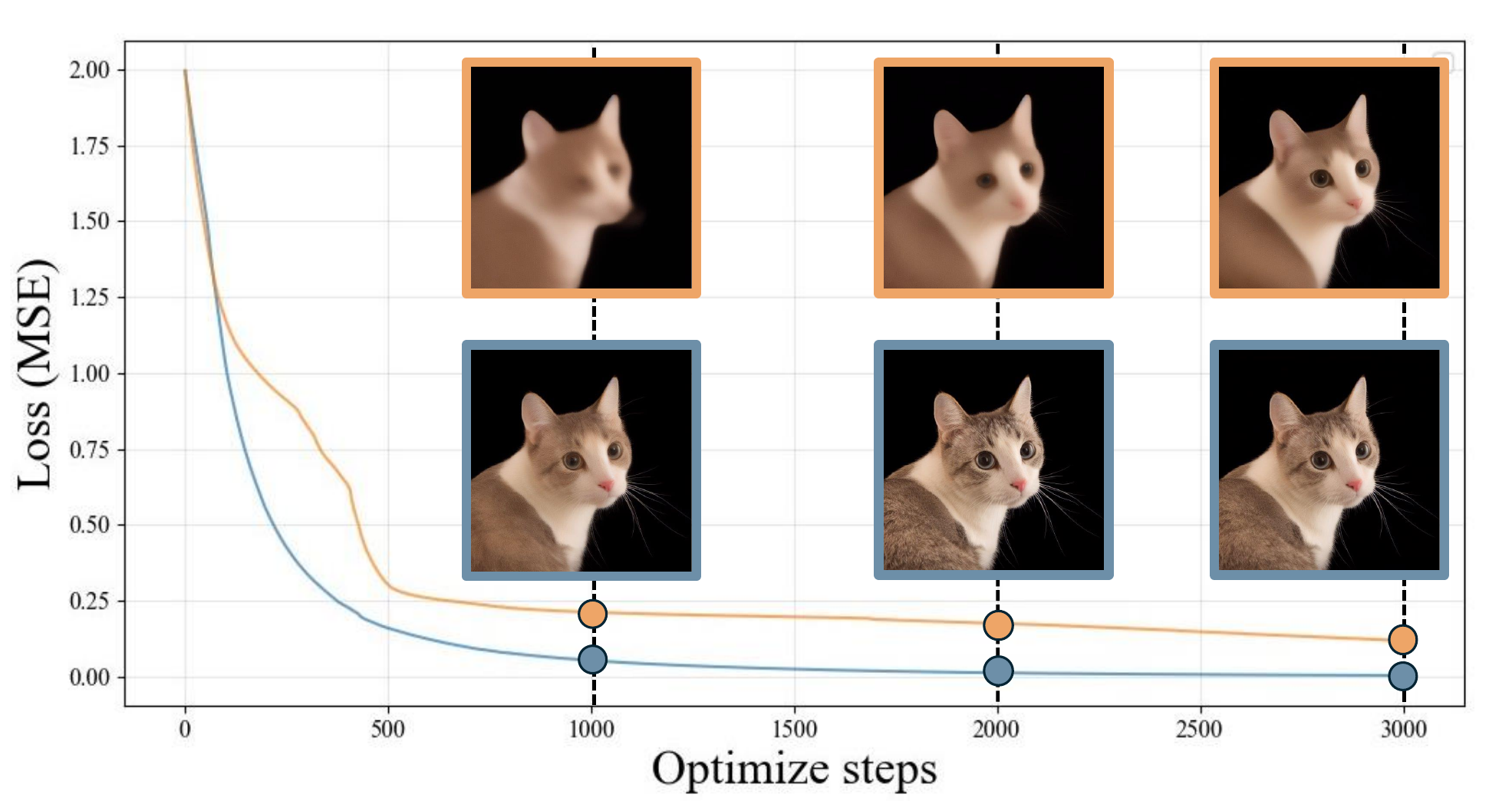}
    \vspace{-10pt}
    \caption{\textbf{Comparison of optimization results using different loss functions.} The orange line represents the optimization process using the full gradient of the MSE loss, while the blue line depicts optimization using MSD loss. The images are sample outputs generated every 1000 optimization steps. The result demonstrates faster convergence but higher image quality when using MSD Loss.}
    \label{fig:loss_graph}


\vspace{-10pt}
\end{figure}

\paragraph{Multistep score distillation.}
Naively applying our method incurs high costs from backpropagating through the denoising network up to $N$ times and requires significant memory usage. Such requirements are one of the main reasons that recent noise optimization work~\cite {eyring2024reno,kim2024model} typically relies on one or few-step models. However, we aim to maximize the number of denoising steps $N$ used in our method since the quality of $\cfgimage$ affects the performance of our model prediction.

To circumvent the backpropagation costs of the full-step diffusion model, we propose a novel approach, multistep score distillation (MSD), inspired by score distillation sampling~\citep{poole2022dreamfusion}, where we detach gradients through a denoising network during backpropagation. 
Specifically, the typical denoising process $\mathcal{L}_{\text{Denoise}}{(g_\phi(x_T),\theta,d)}$ is defined as follows:
 \begin{equation}
     \mathcal{L}_{\text{Denoise}}{(g_\phi(x_T),\theta,d)}: =d\left(D_1\left(\dots D_T(g_\phi(x_T))\right),x_0^{\text{Guide}}\right)
 \end{equation}
 where
 \begin{equation}
     D_t(x) = a_t x_t + b_t \epsilon_\theta^{(t)}(x),
 \end{equation}
and \( a_t \) and \( b_t \) are coefficients derived directly from the DDIM sampler~\citep{song2020denoising} and defined in supplementary material~\ref{supp:preliminaries}.

In MSD, we perform the typical denoising process but detach the gradients on the denoising network \(\epsilon_\theta\) at each step. Specifically: 
\begin{equation}
\mathcal{L}_\text{MSD}(g_{\phi}(x_T), \theta, d) := d\left(F_1\left(\dots F_T(g_\phi(x_T))\right), x_0^{\text{Guide}}\right),
\end{equation}
where
\begin{equation}
F_t(x) = a_t x_t + b_t \, \text{SG}(\epsilon_\theta^{(t)}(x)),
\end{equation}
where $\text{SG}(\cdot)$ denotes the stop-gradient (detach) operation.

Our \ourmodel $g_\phi(\cdot)$ is trained to minimize $\mathcal{L}_\text{MSD}(g_{\phi}(x_T), \theta, d)$. Fig.~\ref{fig:loss_graph}
 compares optimization results with and without detached gradients, showing that disabling gradients in the denoising network leads to faster convergence and sharper images at substantially lower computational costs. We validate our approach, showing that MSD serves as a close approximation to learning with full-gradient objective $\mathcal{L}_{\text{Denoise}}{(g_\phi(x_T),\theta,d)}$. This is clarified in the following proposition. We provide a detailed proof in supplementary material \ref{jacobian_proof}.

\begin{proposition}
By approximating the gradients through Multistep Score Distillation (MSD) using detached gradients at each step, we approximate the full-gradient objective with a mild assumption. In conclusion, the two gradients can be approximated as follows:
\begin{equation}
    \nabla_\phi \mathcal{L}_{\text{Denoise}}(g_\phi(x_T),\theta,d)\approx k\nabla_\phi\mathcal{L}_{\text{MSD}} (g_\phi(x_T),\theta,d),
\end{equation}
where $k\in (0,1)$ is constant.

\end{proposition}

\begin{figure}
    \centering
    \includegraphics[width=\linewidth]{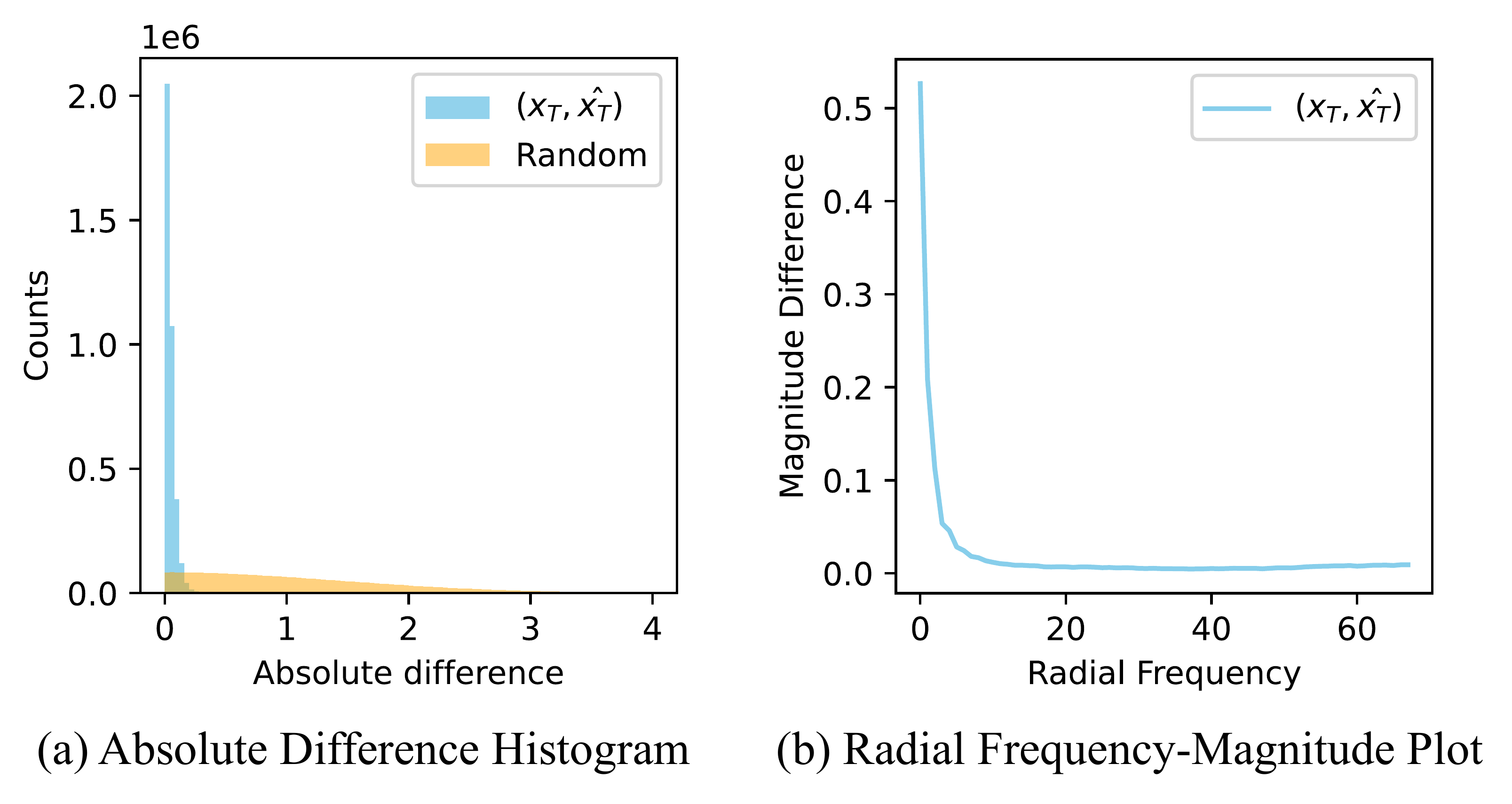}
    \vspace{-20pt}
    \caption{
    \textbf{
    Analysis of the relationship between the initial Gaussian noise \( x_T \) and the refined noise \( \prednoise \).}
    (a) shows a histogram of the absolute difference between \( x_T \) and \( \prednoise \). (b) displays the magnitude difference between the 2D Fourier-transformed frequency components \( \mathcal{F}(x_T) \) and \( \mathcal{F}(\prednoise) \). This demonstrates that the model refines the noise by appropriately adding small, low-frequency components similar to the results shown in Fig.~\ref{fig:analysis}.
    }
    \label{fig:analysis_pred}
\vspace{-10pt}
\end{figure}

 \begin{figure*}[hbt]
    \centering
    \includegraphics[width=1.0\linewidth]{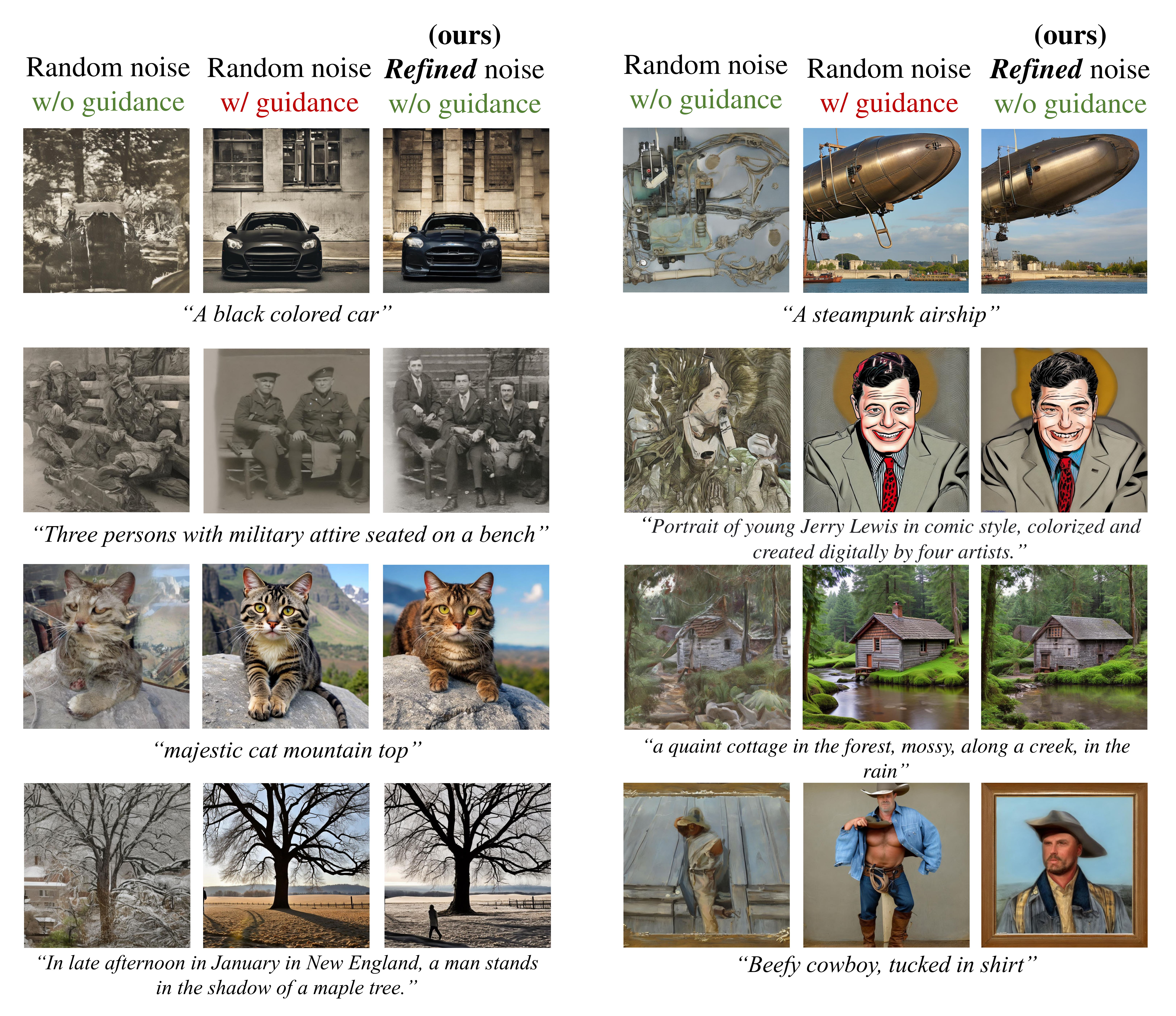}  
    \vspace{-20pt}
    \caption{\textbf{Qualitative results.} Samples starting from Gaussian noise and generated without guidance (left), samples starting from Gaussian noise and generated with sampling guidance (middle), and samples starting from refined noise and generated without guidance (right).}
    \label{fig:main_qual}\vspace{-10pt}
\end{figure*}

\subsection{Dataset Construction}
\label{subsec:asthetic_filtering}
We observe that some proportion of images generated with  CFG~\citep{ho2022classifier} in Stable Diffusion 2.1~\cite{rombach2022high} exhibit low quality, often appearing blurry or displaying distorted facial features, eyes, and noses. Fortunately, our framework is not constrained to CFG. It can incorporate any quality-enhancing techniques applicable at inference time~\citep{hong2024smoothed,hong2023improving}, including PAG~\citep{ahn2024self}, which is known to improve structural accuracy. To enhance the quality of samples, we apply PAG along with CFG, as PAG has been shown to reduce blurriness and improve anatomical structure effectively. Additionally, to mitigate the bias of a fixed guidance scale, we use both CFG and PAG with a randomly varied scale. 

Furthermore, we filter out low-quality images prone to structural issues or artifacts using a human preference model~\citep{schuhmann2022aestheticpredictor}, effectively eliminating poor samples. Detailed implementation details are provided in Sec.~\ref{subsec:datset-filtering}. As shown in Tab.~\ref{table:ablation_table}, this filtering produces superior results compared to training without these enhancements. Note that generating guided image $\cfgimage$ can be done either online or offline; that is, $\cfgimage$ can be pre-generated to enhance computational efficiency.


\begin{table*}[h]
    \centering
    \resizebox{0.8\textwidth}{!}{ 
    \begin{tabular}{c|c|c|ccccc}
        \toprule
        \textbf{Dataset} & \textbf{Initial Noise} & \textbf{Guidance} & \textbf{PickScore~\citep{kirstain2023pick}} & \textbf{HPSv2~\citep{wu2023human}} & \textbf{AES~\citep{schuhmann2022aestheticpredictor}} & \textbf{ImageReward~\citep{xuimagereward}} & \textbf{CLIPScore~\citep{radford2021learning}}
        \\
        \midrule
        \multirow{3}{*}{DrawBench~\citep{saharia2022photorealistic}} 
        & Gaussian & \ding{55} & 19.67 & 0.1689 & 5.129& -1.399 & 25.16
        \\
        & Ours & \ding{55} & \textbf{20.90} & \textbf{0.2306} & \textbf{5.351} & \textbf{-0.291} & \textbf{29.40}
        \\
        \cmidrule(lr){2-8}
        & Gaussian & \ding{51} & 21.29 & 0.2482 & 5.475 & -0.020 & 30.36
        \\
        \midrule
        \multirow{3}{*}{HPD~\citep{wu2023human}} 
        & Gaussian & \ding{55} & 19.18 & 0.1778 & 5.319 & -1.299 & 26.85
        \\
        & Ours & \ding{55} & \textbf{20.47} & \textbf{0.2386} & \textbf{5.608} & \textbf{-0.163} & \textbf{31.21}
        \\
        \cmidrule(lr){2-8}
        & Gaussian & \ding{51} & 21.02 & 0.2469 & 5.788 & 0.159 & 32.27
        \\
        \midrule
        \multirow{3}{*}{Pickapic~\citep{kirstain2023pick}} 
        & Gaussian & \ding{55} & 18.89 & 0.1919 & 5.226 & -1.520 & 24.89
        \\
        & Ours & \ding{55} & \textbf{20.18} & \textbf{0.2347} & \textbf{5.497} & \textbf{-0.304} & \textbf{29.40}
        \\
        \cmidrule(lr){2-8}
        & Gaussian & \ding{51} & 20.67 & 0.2559 & 5.651 & 0.018 & 30.53
        \\
        \midrule
        \multirow{3}{*}{MS-COCO~\cite{lin2014microsoft}} 
        & Gaussian & \ding{55} & 19.56 & 0.1654 & 5.138 & -1.134 & 26.45
        \\
        & Ours & \ding{55} & \textbf{21.01} & \textbf{0.2474} & \textbf{5.368} & \textbf{0.338} & \textbf{30.27}
        \\
        \cmidrule(lr){2-8}
        & Gaussian & \ding{51} &21.59 & 0.2504 &5.487 & -0.058 & 31.29
        \\
        \bottomrule
    \end{tabular}}
    \caption{\textbf{Quantitative comparison of difference metrics across datasets.} Starting from a refined noise using \ourmodel consistently yields higher human preference scores than starting with Gaussian noise, with scores comparable to the guidance case (CFG~\citep{ho2022classifier} + PAG~\citep{ahn2024self}).}
    \label{table:main_quan}
    \vspace{-10pt}
\end{table*}


\section{Experiment}

In this section, to show the effectiveness and efficiency of \ourmodel
, we present extensive qualitative and quantitative results on Stable Diffusion 2.1~\citep{rombach2022high}. We train \ourmodel with text prompts of MS-COCO~\citep{lin2014microsoft} and Pick-a-pic~\citep{kirstain2023pick}. CFG~\citep{ho2022classifier} and PAG~\citep{ahn2024self} are applied to generating images from those datasets with filtering by aesthetic score~\citep{schuhmann2022aestheticpredictor}. We evaluate our model using text prompts of Drawbench~\citep{saharia2022photorealistic}, HPDv2~\citep{wu2023human}, Pick-a-pic~\citep{kirstain2023pick}, and MS-COCO~\citep{lin2014microsoft}. More implementation details and experimental settings can be found in supplementary material~\ref{supp:imple_details}.

\subsection{Results}

\paragraph{Qualitative Comparison.}
We present our qualitative results evaluated on the aforementioned T2I benchmark datasets on Fig.~\ref{fig:main_qual}. We observe notably degraded image quality when the initial noise is Gaussian and guidance is not applied. In contrast, when using the noise refined by our model, we observe consistently superior image quality compared to images from Gaussian noise. Moreover, images from our refined noises without guidance show comparable quality to those from guidance. This result demonstrates the effectiveness of \ourmodel. Additional qualitative results can be found in supplementary materials~\ref{supp:additional_results}.

\vspace{-10pt}

\paragraph{Quantitative Comparison.}

To comprehensively evaluate image fidelity and diversity with FID~\citep{heusel2017gans} and IS~\citep{salimans2016improved}, we generate samples from 30K randomly selected prompts on MS-COCO 2014 validation set~\citep{lin2014microsoft} using three methods: sampling without guidance from Gaussian noise, sampling with guidance from Gaussian noise, and sampling without guidance from noise refined by \ourmodel (ours).

Tab.~\ref{table:main_fid_is} presents FID~\citep{heusel2017gans} and IS~\citep{salimans2016improved} evaluation results and Tab.~\ref{table:main_quan} presents the results of human preference scores~\citep{kirstain2023pick, wu2023human, schuhmann2022aestheticpredictor, xuimagereward} and prompt adherence evaluation~\citep{radford2021learning}. Across all metrics, our model shows a consistent and substantially improved quality over that of images from Gaussian noise, achieving results comparable to those obtained with guidance. Notably, our model achieves a lower FID~\citep{heusel2017gans} score than even the guidance setting, addressing concerns about potential declines in image fidelity and diversity when using model-generated data as training data. This result highlights the advantage of our method in enabling efficient training without relying on large-scale image datasets. More details about those datasets for evaluation can be found in supplementary material~\ref{supp:imple_details}.


\vspace{-10pt}

\paragraph{User Study.}
Tab.~\ref{table:user_study} shows the results of user study, confirming \ourmodel's comparable to results starting from Gaussian initial noise without guidance. 45 participants compared 30 image pairs generated with guidance and our method (refined noise without guidance), using generated images for evaluation in Tab.~\ref{table:main_quan}, and evaluated visual appealiing and prompt alignment.

\begin{table}[h]
    \centering
    \resizebox{0.45\textwidth}{!}{ 
    \begin{tabular}{c|c|cc|c}
        \toprule
        \textbf{Initial Noise} & \textbf{Guidance} & \textbf{FID~\citep{heusel2017gans}} $\downarrow$ & \textbf{IS~\citep{salimans2016improved}} $\uparrow$ & \textbf{Inference Time} $\downarrow$ \\
        \midrule
        Gaussian & \ding{55} & 42.71 & 20.86 & \textbf{1.357s} \\
        Ours & \ding{55} & \textbf{11.39} & \textbf{35.73} & 1.504s \\
        \cmidrule(lr){1-5}
        Gaussian & \ding{51} & 13.38 & 37.64 & 2.589s \\
        \bottomrule
    \end{tabular}}
    \caption{\textbf{Quantitative comparison of image quality and computational cost.} }
    \label{table:main_fid_is}
    \vspace{-5pt}
\end{table}

\begin{table}[h]
    \centering
    \resizebox{0.6\linewidth}{!}{
    \begin{tabular}{c|c|cc}
        \toprule
        \multicolumn{2}{c}{\textbf{Parameter}} & \textbf{FID~\citep{heusel2017gans}} $\downarrow$ & \textbf{IS~\citep{salimans2016improved}} $\uparrow$ \\
        \midrule
        \multirow{2}{*}{\# of steps}
        & 5 & 13.74 & 30.80 \\
        & 10 & \textbf{13.36} & \textbf{32.81} \\
        \midrule
        \multirow{2}{*}{Filtering}
        & \ding{55} & 15.27 & 29.44 \\
        & \ding{51} & \textbf{13.55} & \textbf{31.15} \\
        \bottomrule
    \end{tabular}}
    \caption{\textbf{Ablation study on the number of denoising steps and dataset filtering during training.}}
    \vspace{-5pt}
    \label{table:ablation_table}
\end{table}

\begin{table}[h]
    \centering
    \resizebox{1.0\linewidth}{!}{
    \begin{tabular}{lcc}
        \toprule
        \textbf{Metric} & \textbf{Gaussian Noise + CFG} & \textbf{Refined Noise (Ours)} \\
        \midrule
        Image Quality & 46.04\% & \textbf{53.96\%}   \\
        Prompt Adherence &  48.24\% & \textbf{51.76\%}  \\
        \bottomrule 
    \end{tabular}
    }
    \caption{\textbf{User study on the image quality and prompt adherence of generated images.}}
    \label{table:user_study}
\end{table}

\subsection{Ablation Study}

\paragraph{Number of denoising steps.}
We demonstrate that the number of denoising steps significantly impacts performance. Specifically, we compare cases with denoising steps \( N=5 \) and \( N=10 \), reporting FID~\citep{heusel2017gans}, IS~\cite{salimans2016improved} in Tab.~\ref{table:ablation_table}. The results indicate improved performance with a bigger number of denoising steps. Considering that a high number of steps (e.g., \( N \geq 10 \)) incurs prohibitive backpropagation costs, this supports the necessity of MSD to circumvent backpropagation costs.

\paragrapht{Dataset filtering.}
\label{subsec:datset-filtering}
To evaluate the impact of dataset filtering, we generate images using 80K prompts from the MS-COCO~\citep{lin2014microsoft} dataset and remove those with an aesthetic score (AES)~\citep{schuhmann2022aestheticpredictor} below 6.0, resulting in about 25\% of the images remaining. As shown in Tab~\ref{table:ablation_table}, metrics for generated images are significantly improved in the filtered dataset, demonstrating the substantial impact of dataset filtering on enhancing guidance-free image generation.

\section{Analysis and Discussion}

We analyze what \ourmodel learns and identify components in refined noise that contribute to guidance-free generation, discussing the advantages of working in this space.

\subsection{What Does \ourmodel Learn?}
In Fig.~\ref{fig:analysis_pred}, we show that our model refines the noise by adding mostly small, low-frequency components. The distribution of absolute norm and frequency of the added noise difference is similar to the noise difference between a Gaussian noise $x_T$ and the inversion noise $\invnoise$ shown in Fig.~\ref{fig:analysis}, without explicit constraints to achieve this.

\begin{figure}
    \centering
    \includegraphics[width=1.0\linewidth]{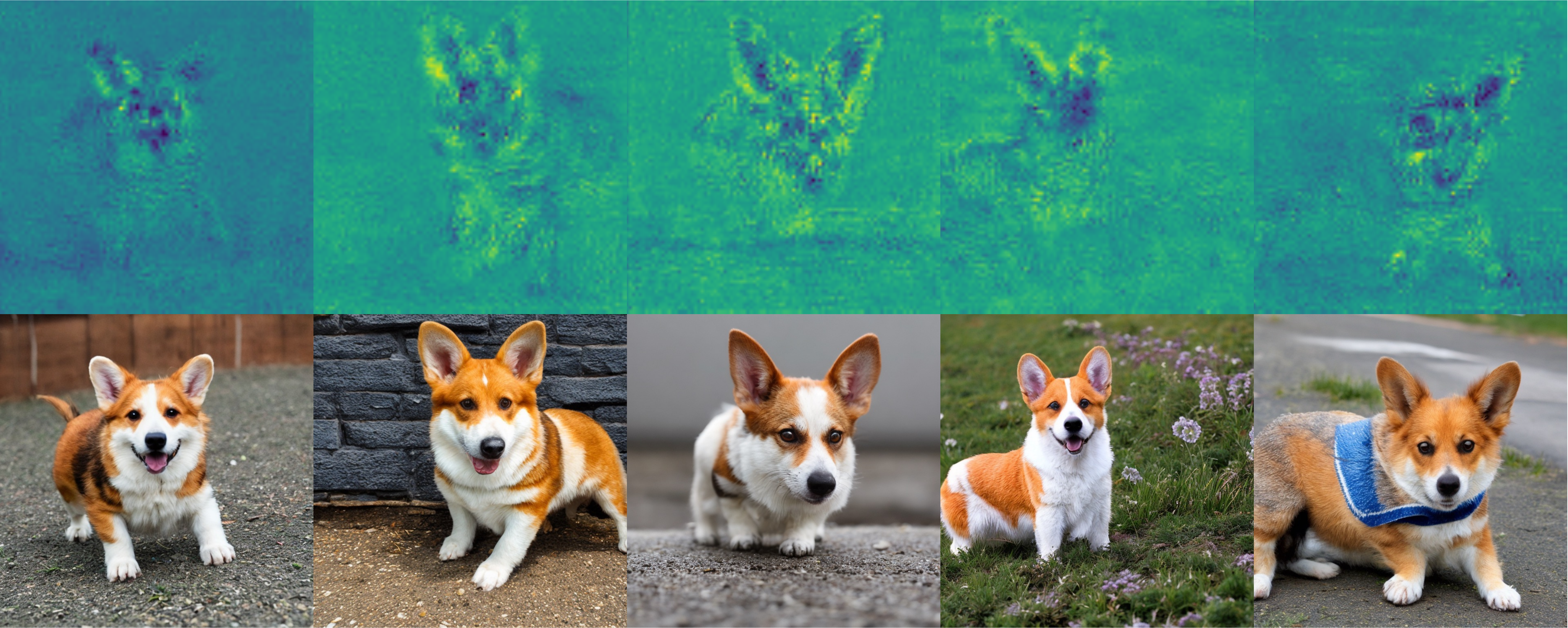}
    \caption{\textbf{Visualization of noise difference between \( x_T \) and \( \prednoise \).}
    The top row shows the difference, while the bottom row displays the corresponding generated images. The added signal functions as a layout, guiding the structure of the image during generation. Here, prompt \textit{`a photo of corgi'} is used.
    }
    \label{fig:noise_delta}
\vspace{-10pt}
    
\end{figure}

\begin{figure}
    \centering
    \includegraphics[width=1.0\linewidth]{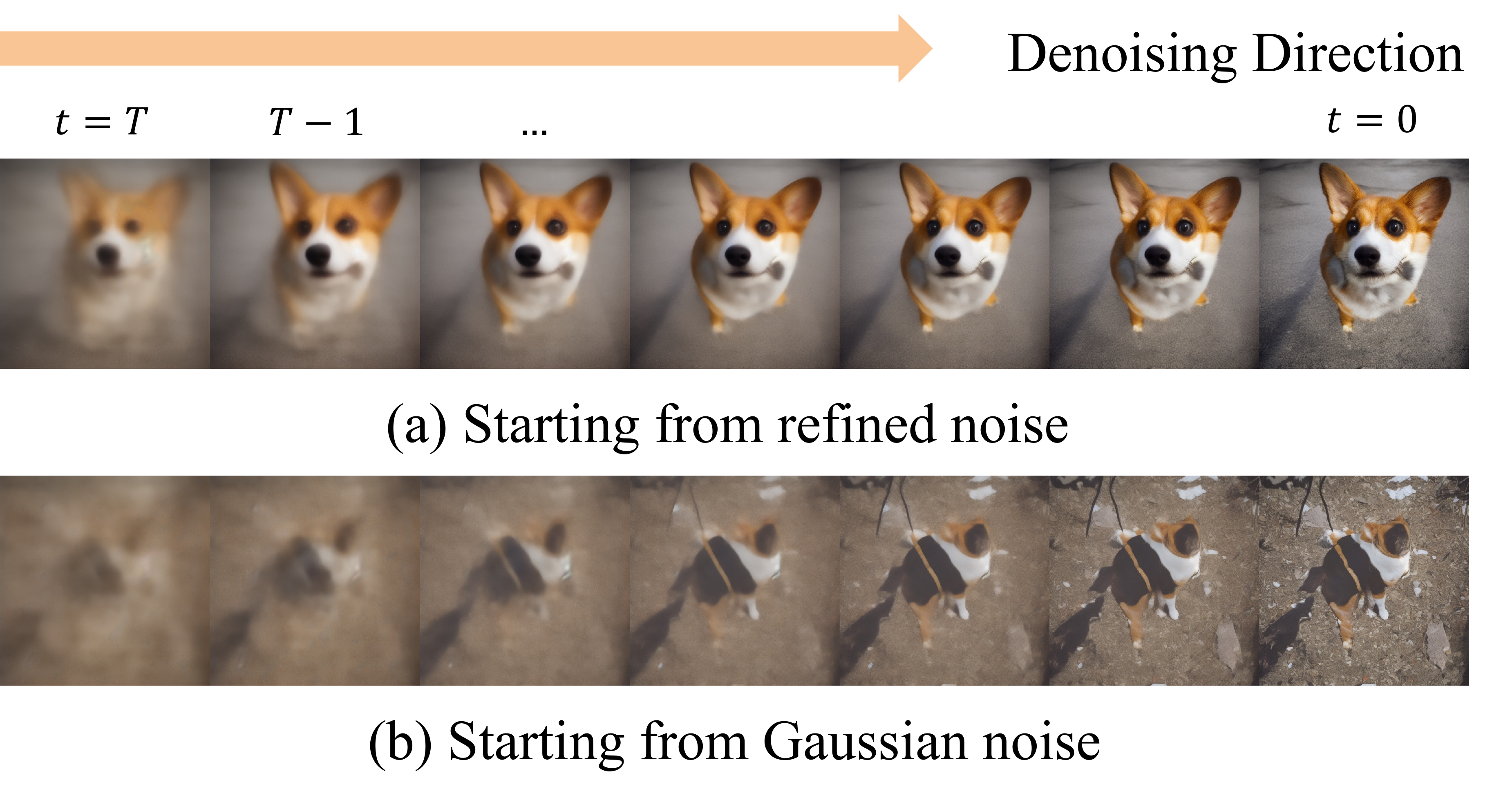}
    \vspace{-18pt}
    \caption{\textbf{Visualization of denoising process.}
    (a) Starting from refined noise aids the model in establishing the overall layout early in the generation process, facilitating the successful creation. (b) In contrast, when beginning with Gaussian noise, the model struggles to capture the overall layout, resulting in incomplete or disjointed details rather than producing a fully plausible image.
    }
    \label{fig:noise_denoising_process}
    \vspace{-10pt}
\end{figure}

\paragrapht{Low-frequency components aid denoising.}
Fig.~\ref{fig:noise_delta} shows that the low-frequency components in the noise difference are dependent on the condition (e.g., text prompt) and act as an initial \textit{layout} for the synthesized image. These low-frequency signals significantly help diffusion models in forming object shapes in the early steps of denoising. Fig.~\ref{fig:noise_denoising_process} (a) shows that starting from refined noise, the model quickly establishes plausible images at much earlier stages, allowing the model to focus on adding details within the given layout during denoising. In contrast, as shown in Fig.~\ref{fig:noise_denoising_process} (b), the diffusion model struggles to create a coherent layout in the early denoising steps, resulting in partial details filled in incorrect locations, often leaving ambiguous regions untouched throughout the denoising process. 
\vspace{-5pt}
\paragrapht{Diversity and generalizability.} 
While these initial layouts might appear to limit diversity, Fig.~\ref{fig:noise_delta} shows that the generated layouts vary significantly depending on the initial noise. We confirm this with diversity metrics of IS~\citep{salimans2016improved}, demonstrating greater diversity than guidance-based methods.
 In addition, our model generalizes beyond the training data, performing well with unseen noise and prompts (Fig.~\ref{fig:noise_delta}, Table.~\ref{table:main_quan}), suggesting a generalizable mapping between initial and refined noise.

\begin{figure}
    \centering
    \includegraphics[width=1.0\linewidth]{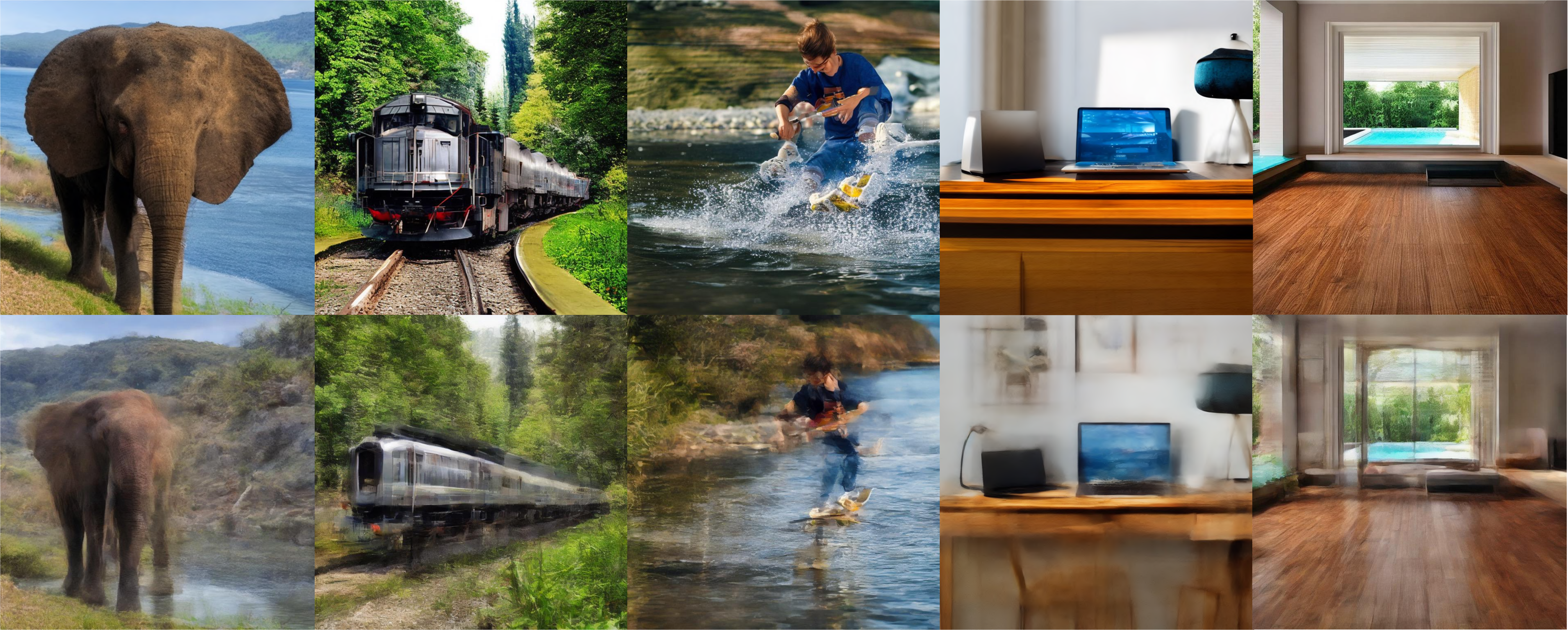}
    \caption{\textbf{Comparison of training \ourmodel $g_\phi$ (top) and the denoising network $\epsilon_\theta$ (bottom).} This demonstrates that tuning the input of the diffusion pipeline converges faster than tuning the pipeline (denoising network) within the same number of training iterations.}
    \label{fig:model-tuning}
    \vspace{-10pt}
\end{figure}

\subsection{Why Learn Noise Mapping?}
We consider the rise of prompt learning. Large-scale models like CLIP~\citep{radford2021learning}, trained on web-scale datasets often contain up to billions of parameters. Fine-tuning such models is impractical and risks disturbing well-learned representations~\citep{zhou2022learning}. Instead, tuning the input prompts of the model, a method known as \textit{prompt learning}, has gained popularity as an effective approach~\citep{zhou2022learning,zhou2022conditional,jiang2020can,shin2020autoprompt}.

In this context, limiting training to the noise space can be seen as an efficient alternative to tuning the entire denoising pipeline. As seen in Fig.~\ref{fig:analysis_pred} and Fig.~\ref{fig:noise_delta}, certain low-frequency components in the noise space provide critical information, such as image layout, which allows learning with a smaller dataset without the need to tune all model parameters. Unlike typical text-to-image diffusion models~\citep{rombach2022high} or their distilled versions~\citep{sauer2025adversarial,lin2024sdxl,podell2023sdxl}, which require up to billions of images~\citep{schuhmann2022laion}, our model achieves effective training with only 50K noise and self-generated image pairs.

We verify this by tuning the diffusion model. However, as shown in Fig.~\ref{fig:model-tuning}, this approach led to slower convergence and frequent loss explosions, making training unstable. Details and discussion of this comparison can be found in supplementary material~\ref{supp:discussion}.

\section{Conclusion}

In this work, we propose \textbf{\ours}, an efficient and effective approach to replacing guidance in diffusion sampling with a noise refinement by a single neural network forward pass. Our noise refining model functions as a plug-and-play module based on the original diffusion model and significantly improves image fidelity. Furthermore, our method is highly efficient, which can be trained using lightweight lora, requiring only a small set of model-generated images for training and remaining feasible on consumer-grade GPUs, thanks to our proposed MSD loss. Beyond its practicality, we believe this work serves as a stepping stone toward a deeper understanding of the role of guidance and noise in diffusion models.

{
    \small
    \bibliographystyle{ieeenat_fullname}
    \bibliography{main}
}
\appendix
\onecolumn

\begin{center}
    \textbf{\Large A Noise is Worth Diffusion Guidance}
    \vspace{1em} \\
    \textbf{\large Supplementary Material}
\end{center}
\vspace{0.5em}

\setcounter{page}{1}

In the supplementary material, we clarify the notations and formulations related to diffusion models used in the main paper and provide the proofs for our propositions (Section~\ref{supp:theory}), more ablation studies regarding \ourmodel (Section~\ref{supp:more_ablation}), additional results including qualitative results, comparison with other methods, user study (Section~\ref{supp:additional_results}), implementation details and experimental settings (Section~\ref{supp:imple_details}) and further discussions (Section~\ref{supp:discussion}).

\section{Theoretical Background}
\label{supp:theory}
\subsection{Preliminaries}
\label{supp:preliminaries}

\paragraph{Denoising Diffusion Probabilistic Models (DDPM).}
DDPM~\cite{ho2020denoising} defines a forward process that derives $x_t$ by adding Gaussian noise to the image $x_{t-1}$ according to the variance schedule, and a reverse process that samples $x_{t-1}$ from $x_t$, both as a Markovian chain. The forward process is defined as
\begin{align}
    q(x_t | x_{t-1}) &= \mathcal{N}\left(x_t; \sqrt{\frac{\alpha_t}{\alpha_{t-1}}} \, x_{t-1}, \left(1 - \frac{\alpha_t}{\alpha_{t-1}}\right) \mathbf{I}\right), \\
    q(x_t | x_0) &= \mathcal{N}(x_t; \sqrt{\alpha_t} \, x_0, (1 - \alpha_t) \mathbf{I}),
\end{align}
with noise rate at timestep $t$ as $1 - {\alpha_t}/{\alpha_{t-1}}$, where $\alpha_t$ denotes noise scaling factors up to time step $t$. The reverse process is defined below.
\begin{align}
    p_\theta(x_{t-1} | x_t) &= \mathcal{N}\left(x_{t-1}; \mu_\theta^{(t)}(x_t), \sigma_t^2 \mathbf{I}\right).
\end{align}
To reparameterize the equation using
\begin{align}
    x_t &= \sqrt{\alpha_t}x_0 + \sqrt{1 - \alpha_t} \, \epsilon \quad \text{for } \epsilon \sim \mathcal{N}(0,\mathbf{I}),
\end{align}
and $\epsilon_{\theta}$, which is a function approximator for predicting $\epsilon$ from $x_t$, the inference process becomes
\begin{align}
    x_{t-1} &= \frac{1}{\sqrt{\frac{\alpha_t}{\alpha_{t-1}}}} \left( x_t - \frac{1 - \frac{\alpha_t}{\alpha_{t-1}}}{\sqrt{1 - \alpha_t}} \, \epsilon_\theta^{(t)}(x_t) \right) + \sigma_t z,
\end{align}
Where $z \sim \mathcal{N}(0,\mathbf{I}) $ and $\sigma_t^2$ denotes the variance of Gaussian trainsitions  .The objective of DDPM is defined as 
\begin{align}
    L_{\text{simple}}(\theta) &= \mathbb{E}_{t, x_0, \epsilon} \left[ \| \epsilon - \epsilon_\theta^{(t)}(x_t) \|^2 \right],
\end{align}
where the L2 loss between the actual noise $\epsilon$ added during training and the noise prediction $\epsilon_{\theta}(x_t,t)$ for uniformly sampled $t\in \{ 1,...,T \}$.
\paragraph{Denoising Diffusion Implicit Models (DDIM).}
DDIM~\cite{song2020denoising} consider the following inference distributions:
\begin{align}
    q_\sigma(x_{1:T} | x_0) &:= q_\sigma(x_T | x_0) \prod_{t=2}^T q_\sigma(x_{t-1} | x_t, x_0).
\end{align}
with a mean function as below.
\begin{equation}
    q_\sigma(x_{t-1} | x_t, x_0) = \mathcal{N} \left( \sqrt{\alpha_{t-1}} x_0 + \sqrt{1 - \alpha_{t-1} - \sigma_t^2} \cdot \frac{x_t - \sqrt{\alpha_t} x_0}{\sqrt{1 - \alpha_t}}, \sigma_t^2 \mathbf{I} \right).
\end{equation}
Distinctively from DDPM, the forward process is Non-Markovian since each $x_t$ could depend on both $x_{t-1}$ and $x_0$. Reparameterizing with $\epsilon_{\theta}$, we can sample $x_{t-1}$ from $x_t$ through an equation:
\begin{equation} \label{eq:ddim_reverse_step}
     \begin{split}
x_{t-1} &= \sqrt{\alpha_{t-1}} \underbrace{\left( \frac{x_t - \sqrt{1 - \alpha_t} \, \epsilon_\theta^{(t)}(x_t)}{\sqrt{\alpha_t}} \right)}_{\text{predicted } x_0} + \sqrt{1 - \alpha_{t-1} - \sigma_t^2} \cdot \epsilon_\theta^{(t)}(x_t) + \sigma_t \epsilon_t \nonumber\\
&=  a_t x_t + b_t \epsilon_\theta^{(t)}(x),
\end{split}
\end{equation}
where $\epsilon_{t} \sim \mathcal{N}(0,\mathbf{I})$ and
$a_t = \sqrt{{\alpha_{t-1}}}/\sqrt{{\alpha_t}}, \quad
b_t = \sqrt{1 - \alpha_{t-1}} - a_t\sqrt{1-\alpha_t}.$

The objective of DDIM is the same as that of DDPM:
\begin{equation}
    L_{\text{DDIM}}(\theta) = \mathbb{E}_{t, x_0, \epsilon} \left[ \| \epsilon - \epsilon^{(t)}_\theta(x_t) \|^2 \right].
\end{equation}

\paragraph{Denoising and inversion process.}

We denote the denoising process as $\text{Denoise}(x_T)$. When using the DDIM sampler~\citep{song2020denoising}, the denoising process is defined as:
\begin{equation}
\label{eq:denoise}
    \text{Denoise}(x_T) := D_1\left(\dots D_T(g_\phi(x_T))\right),
\end{equation}
where each step $D_t$ is given by:
\begin{equation}
\label{eq:d_t(x)}
    D_t(x) := a_t x_t + b_t \epsilon_\theta^{(t)}(x).
\end{equation}

The guided denoising process, denoted as $\text{Denoise}^{\text{Guide}}(x_T, c)$, follows the same steps as Eq.~\ref{eq:denoise}, but replaces $\epsilon_\theta^{(t)}(x)$ with guided scores, such as the classifier-free guided score $\epsilon^{\text{CFG}}_{\theta}(x_t, c)$~\citep{ho2022classifier}, the perturbed-attention guided score $\epsilon^{\text{PAG}}_{\theta}(x_t)$~\citep{ahn2024self}, or a combination of both ($\epsilon^{\text{CFG,PAG}}_{\theta}(x_t)$). These guided scores are defined in Eqs.~\ref{eq:cfg},~\ref{eq:pag}, and~\ref{eq:cfgpag}.

While we utilize the DDIM scheduler in this work, any other diffusion scheduler~\citep{ho2020denoising,song2020denoising,karras2022elucidating} can be used by appropriately modifying $a_t$ and $b_t$.

For the inversion process $\text{Inversion}(x_0, c)$, we follow the method in~\citep{garibi2024renoise} to obtain the initial noise $x_T$, which can be denoised back to the given image $x_0$ without employing any guidance methods~\citep{ho2022classifier,ahn2024self} during inversion.

\subsection{Derivations}
\label{supp:derivations}
\paragraph{Proposition 1.}
Let $x_T$ be an initial noise, and suppose that $x_0$ is the image obtained through denoising. Assuming Lipschitz continuity with distance metric $d$, for every $x_T$, there exists a constant $\kappa>0$ such that the following holds:
\[d(x_T,\trueinvnoise)<\kappa d(x_0,\cfgimage).\]

\noindent\textbf{\textit{proofs.}}
The Lipschitz condition is expressed as follows:
\begin{equation}
\label{lipschitz}
      d(\epsilon_\theta^{(t)}(x),\epsilon_\theta^{(t)}(y))\leq{L_t} d (x,y),
\end{equation}
where $L_t$ is constant dependent on $t$, $x$ and $y$ are arbitrary inputs to \(\epsilon_\theta^{(t)}\).
DDIM step in terms of $x_t$ can be expressed as follows:
    \begin{equation}
        x_{t-1}=\sqrt{\frac{\alpha_{t-1}}{\alpha_{t}}}x_t+\left(\sqrt{1-\alpha_{t-1}}-\sqrt{\frac{\alpha_{t-1}(1-\alpha_t)}{\alpha_t}}\right)\epsilon_\theta^{(t)}(x_t).
    \label{1step_ddim}
    \end{equation} 
Eq.~\eqref{1step_ddim} can be expressed in terms of $x_t^{\text{Guide}\dagger}$ which is denoised from $\trueinvnoise.$ With those equations, we can get the following equation,
    \begin{equation}
     \begin{split}
    x_{t-1} - x_{t-1}^{\text{Guide}\dagger} &=
    \sqrt{\frac{\alpha_{t-1}}{\alpha_t}}(x_t - x_t^{\text{Guide}\dagger})+
    \left(\sqrt{1-\alpha_{t-1}} - \sqrt{\frac{\alpha_{t-1}(1-\alpha_t)}{\alpha_t}}\right)
    (\epsilon_\theta^{(t)}(x_t) - \epsilon_\theta^{(t)}(x_t^{\text{Guide}\dagger}))\nonumber\\
    &= \sqrt{\frac{\alpha_{t-1}}{\alpha_t}}(x_t - x_t^{\text{Guide}\dagger})-\gamma_t(\epsilon_\theta^{(t)}(x_t) - \epsilon_\theta^{(t)}(x_t^{\text{Guide}\dagger})),
    \end{split}   
    \end{equation}
    where $\gamma_t=\left(\sqrt{\alpha_{t-1}(1-\alpha_{t})/{\alpha_{t}}}-\sqrt{1-\alpha_{t-1}}\right)>0$. If the distance metric $d$ have translation invariance, the equation can be expressed as follows with Eq.~\eqref{lipschitz}:
    \begin{equation}
        d(x_{t-1},x_{t-1}^{\text{Guide}\dagger})
        \leq \sqrt{\frac{\alpha_{t}}{\alpha_{t-1}}}(1+\gamma_t L_t) d(x_{t},x_{t}^{\text{Guide}\dagger})
        \label{1step}.
    \end{equation}
    Recursively organizing Eq.~\eqref{1step} for \( t = T, T-1, \ldots, 1 \), it can be expressed as follows:
    \begin{equation}
         d(x_T,x_T^{\text{Guide}\dagger})\\
        \leq \left(\prod_{t=1}^{T}(1+\gamma_t L_t)\right)\sqrt{\frac{\alpha_{T}}{\alpha_0}} d(x_{0},x_{0}^{\text{Guide}\dagger}).\\
    \end{equation}
    Since $\alpha_T$ is close to $0$, using $d(x_0,x_0^{\text{Guide}\dagger})$ is sufficient to directly learn $x_T^{\text{Guide}\dagger}$ if $d(x_{0},x_{0}^{\text{Guide}\dagger})$ is small enough.
\paragraph{Proposition 2.}
\label{jacobian_proof}
By approximating the gradients through Multistep Score Distillation (MSD) using detached gradients at each step, we approximate the full-gradient objective with a mild assumption. In conclusion, the two gradients can be approximated as follows:
\begin{equation}
    \nabla_\phi \mathcal{L}_{\text{Denoise}}(g_\phi(x_T),\theta,d)\approx k\nabla_\phi\mathcal{L}_{\text{MSD}} (g_\phi(x_T),\theta,d),
\end{equation}
where $k\in (0,1)$ is constant.\\

\noindent\textbf{\textit{proofs.}}
Since the only difference between the two losses is the stop gradient in the diffusion model and all other components are identical, it suffices, by the chain rule, to show that the gradient of $F_1(F_2(\dots F_T(g_\phi(x_T))$ with respect to $\phi$ is proportional to the gradient of $\text{Denoise}(g_\phi(x_T))$ with respect to $\phi$. The derivation proceeds as follows:
\begin{equation}
\begin{split}
    \nabla_\phi \text{Denoise}(g_\phi(x_T))&= \nabla_\phi \left(\sqrt{\frac{\alpha_{0}}{\alpha_T}}g_\phi(x_T)-\sum_{t=1}^{T} \sqrt{\frac{\alpha_{0}}{\alpha_{t-1}}} \gamma_t \epsilon_\theta^{(t)}(x_k)\right)\\
&=\left(\sqrt{\frac{\alpha_{0}}{\alpha_T}}I-\sum_{t=1}^{T}\gamma_t \sqrt{\frac{\alpha_{0}}{\alpha_{t-1}}}\frac{\partial\epsilon_\theta^{(t)}(x_t)}{\partial x_t}\frac{\partial x_t}{\partial g_\phi(x_T)}\right)\frac{\partial g_{\phi}(x_T)}{\partial \phi}.
\label{gradient}
\end{split}
\end{equation}
As detailed in~\ref{supp:more_ablation}, the term ${\partial\epsilon_\theta^{(t)}(x_k)}/{\partial x_k}$ can be approximated as being proportional to the identity matrix. Additionally, the term ${\partial x_k}/{\partial g_\phi(x_T)}$ can be expressed in terms of ${\partial\epsilon_\theta^{(t)}(x_k)}/{\partial x_k}$. Then, each component of ${\partial\epsilon_\theta^{(t)}(x_k)}/{\partial x_k}$ can be approximated by the identity matrix. Consequently, $({\partial\epsilon_\theta^{(t)}(x_k)}/{\partial x_k})\left({\partial x_k}/{\partial g_\phi(x_T)}\right)$ becomes proportional to the identity matrix. Denoting the proportionality constant as $\eta_t$, Eq.~\eqref{gradient} is simplified as follows:
\begin{equation}
\begin{split}
    \eqref{gradient}&=\left(\sqrt{\frac{\alpha_{0}}{\alpha_T}}-\sum_{t=1}^{T}\sqrt{\frac{\alpha_{0}}{\alpha_{t-1}}}\gamma_t \eta_t\right)\frac{\partial g_{\phi}(x_T)}{\partial \phi}\\
    &=\left(1-\sqrt{\alpha_T}\sum_{t=1}^{T}\frac{1}{\sqrt{\alpha_{t-1}}}\gamma_t \eta_t\right)\sqrt{\frac{\alpha_{0}}{\alpha_T}}\frac{\partial g_{\phi}(x_T)}{\partial \phi}\\
    &=\left(1-\sqrt{\alpha_T}\sum_{t=1}^{T}\frac{1}{\sqrt{\alpha_{t-1}}}\gamma_t \eta_t\right)\nabla_\phi F_1(F_2(\dots F_T(g_\phi(x_T))).
\end{split}
\end{equation}\\

\clearpage
\section{More Ablation Studies}

\label{supp:more_ablation}

\subsection{Diffusion Model Jacobian Approximation}
\begin{figure}[!h]
    \centering
    \includegraphics[width=0.7\linewidth]{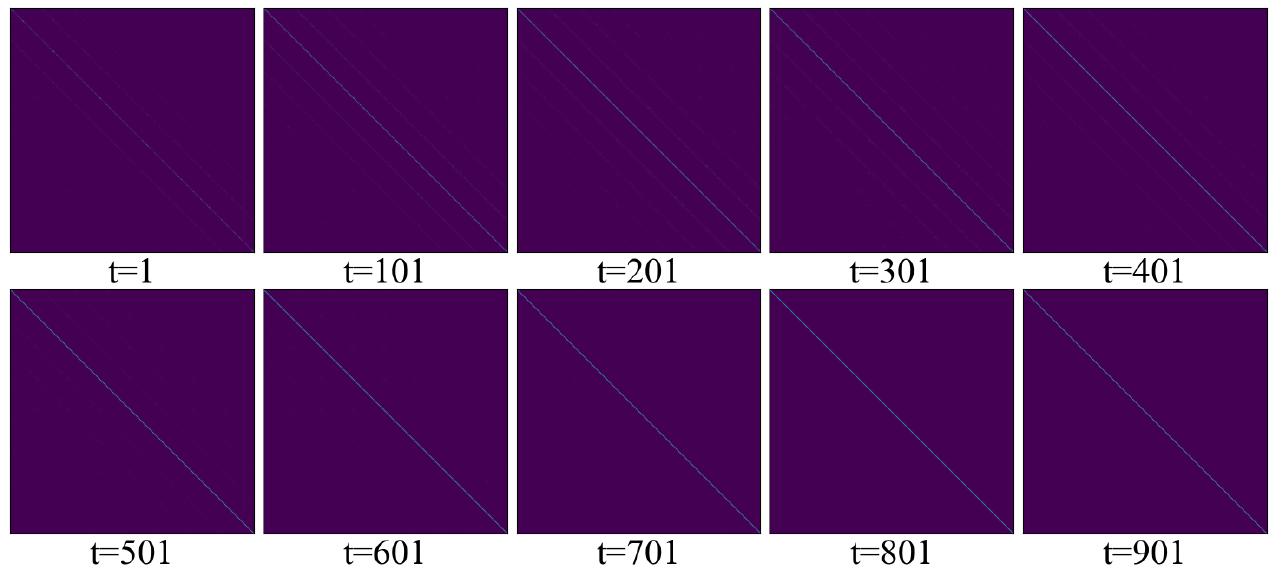}
    \caption{\textbf{Visualization of Jacobian of a denoising network.} Starting from $T=1000$, we performed denoising over 10 steps and plotted the Jacobian heatmap at each timestep.We extracted a $500\times 500$ section from the full Jacobian matrix for visualization. Each plot demonstrates that the Jacobian is close to the identity matrix.}
    \label{fig:unet_jacobian} 
\vspace{-100pt}
\end{figure}

\begin{figure}
    \centering
    \includegraphics[width=0.5\linewidth]{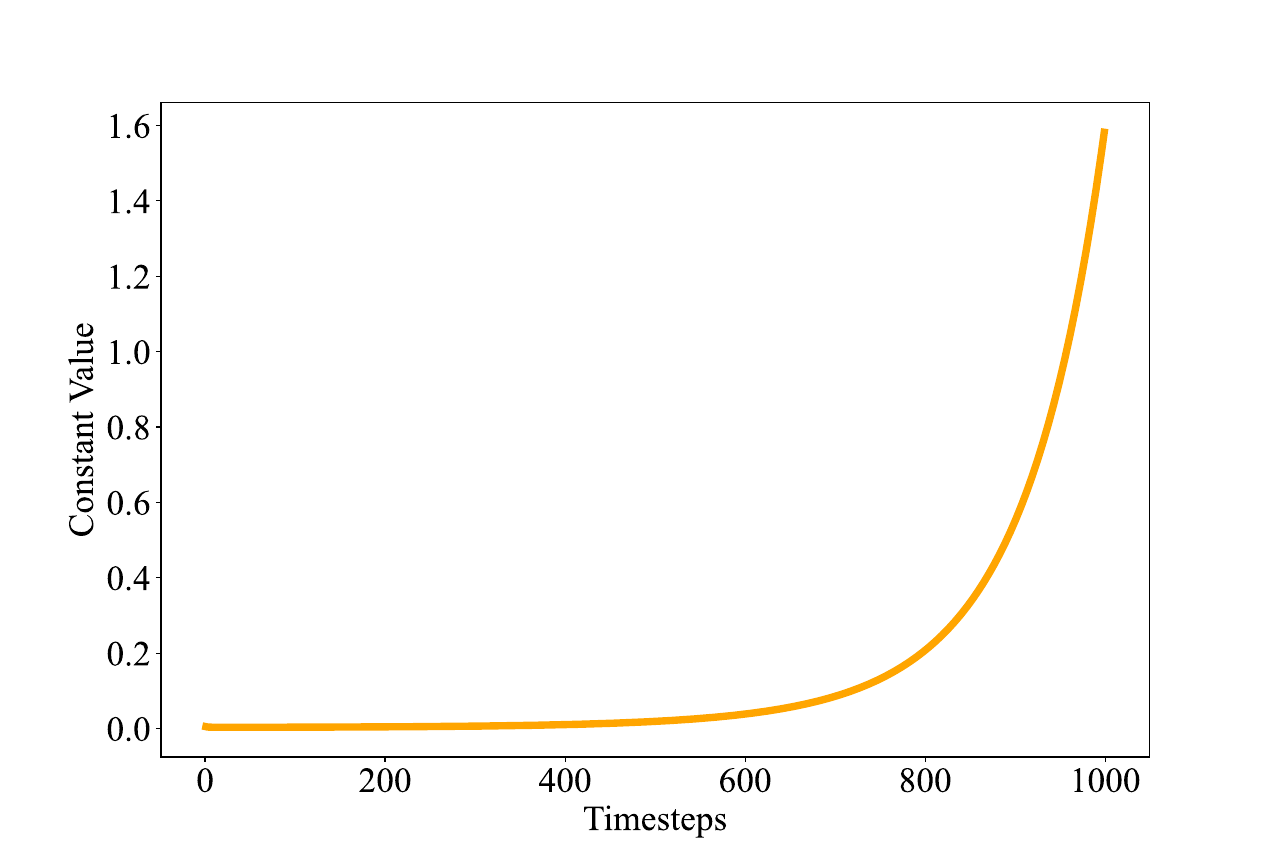}
    \caption{\textbf{Visualization of constant values $\frac{\gamma_t}{\sqrt{\alpha_{t-1}}}$ over timesteps.}  
 Visualization of the time-dependent constant value $\frac{\gamma_t}{\sqrt{\alpha_{t-1}}}$ corresponding to Equation (19) across different timesteps. The results numerically demonstrate that for small timesteps, where the Jacobian deviates from the identity matrix, the multiplied constant values are sufficiently close to zero.}
    \label{fig:constant_graph}
\end{figure}

In this section, we present experimental results demonstrating that the Jacobian of the diffusion model $\epsilon^t_\theta$ with respect to the input $x_t$ can be approximated as proportional to the identity matrix. Fig.~\ref{fig:unet_jacobian} illustrates the Jacobian of $\epsilon^t_\theta$. We observe that the  Jacobian of diffusion model behaves like the identity regardless of the timestep, except when
t is significantly small. In such cases, the deviation does not affect our primary analysis. According to the results of \textbf{Proposition 2}, the timestep-dependent constant multiplied to each Jacobian term $\eta_t$ is expressed as follows:

    \begin{equation}
        \frac{1}{\sqrt{\alpha_{t-1}}}\gamma_t=\sqrt{\frac{1-\alpha_t}{\alpha_t}}-\sqrt{\frac{1-\alpha_{t-1}}{\alpha_{t-1}}}.
    \end{equation}
This value can be numerically determined based on the scheduling, and in the case of DDIM~\citep{song2020denoising}, it is presented in Fig.~\ref{fig:constant_graph}. The graph shows that the constant decreases as t approaches 0, becoming 0.
\vspace{-5pt}


%

\clearpage
\subsection{Utilizing Pretrained Knowledge of Diffusion Models}
\label{sec:supp:pretrained}
To train \ourmodel, we adopt attaching LoRA layers to the original model to effectively leverage its pretrained knowledge. To assess the impact of pretrained knowledge, we conduct an ablation study. The comparison involves \ourmodel and the same UNet architecture of the original model, but trained from scratch. We used only filtered MS COCO dataset among both datasets and trained models for 25K steps using two RTX 3090 GPUs. All the other experimental settings are kept consistent.

\begin{figure}[H]
    \centering
    \begin{minipage}{0.55\textwidth} 
        \centering
        \includegraphics[width=1.0\linewidth]{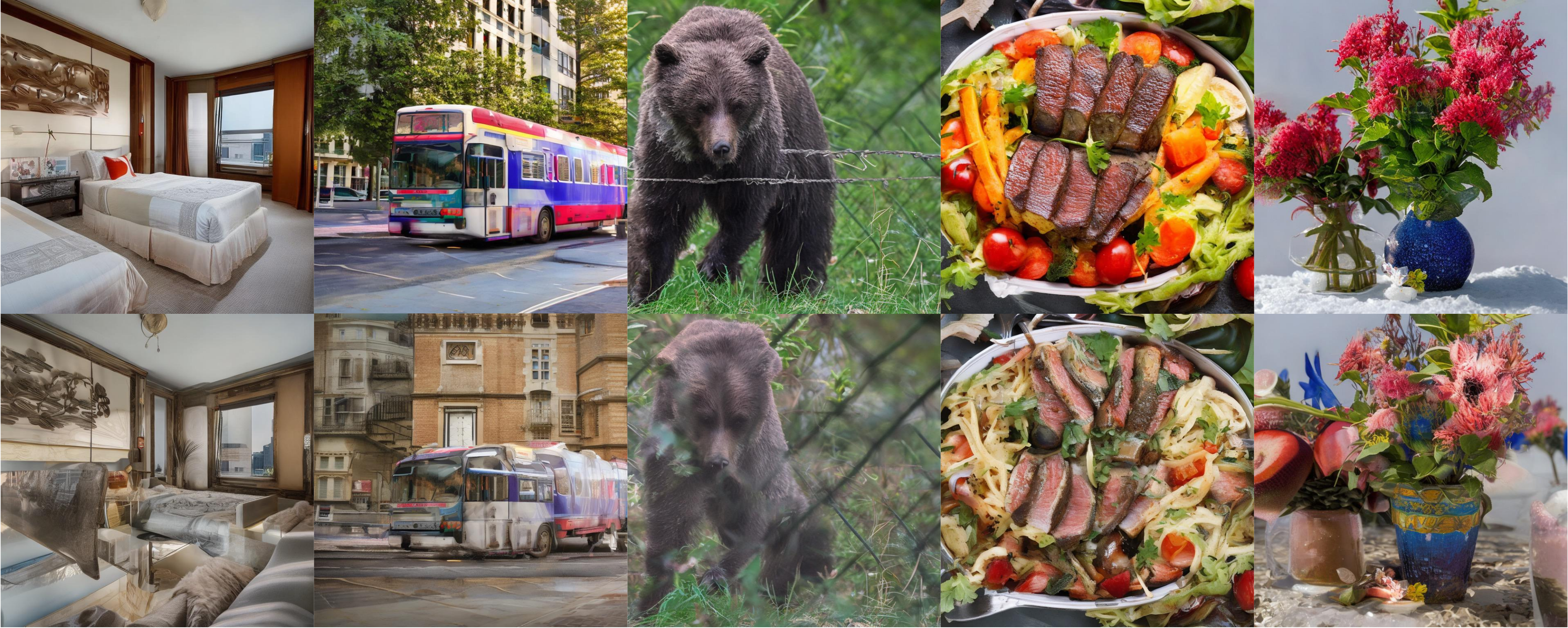}
        \caption{
        \textbf{Qualitative comparison with \ourmodel (top) and UNet trained from scratch (bottom).}
        }
        \label{supp:qual_scratch}
    \end{minipage}
    \hfill 
    \begin{minipage}{0.4\textwidth} 
    \centering
    \begin{tabular}{ l | c }
        \toprule
            \textbf{Model} & \textbf{FID} \\
        \midrule
             UNet trained from scratch & 37.87 \\
             Pretrained + LoRA (\ourmodel) &  13.74 \\
        \bottomrule
    \end{tabular}
    \captionof{table}{
    \textbf{Quantitative comparison with \ourmodel and UNet trained from scratch.}
    }
    \label{supp:quan_scratch}
    \end{minipage}
\end{figure}

Qualitative results are presented in Fig.~\ref{supp:qual_scratch}, and quantitative results are detailed in Tab.~\ref{supp:quan_scratch}. Both results indicate that leveraging pretrained knowledge results in superior performance compared to training from scratch. 

\clearpage

\section{Additional Results}
\label{supp:additional_results}
\subsection{Additional Qualitative Results}
\begin{figure}[!b]
    \centering
    \includegraphics[height=1.0\textheight]{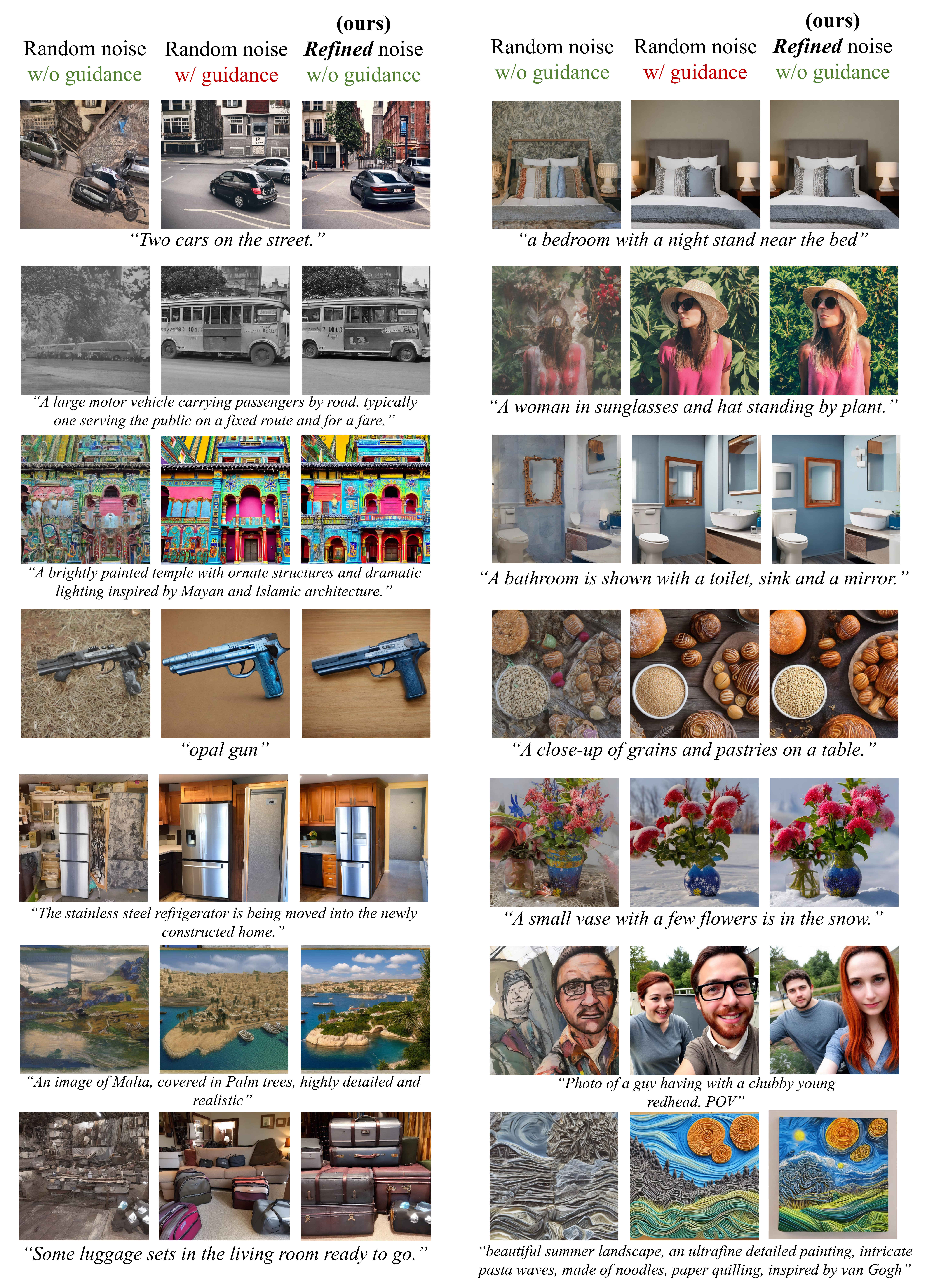}
    \vspace{-10pt}
    \caption{
    \textbf{Additional qualitative results.}
    }
    \label{fig:supp_qual}
\end{figure}

\begin{figure}[!b]
    \centering
    \includegraphics[height=1.0\textheight]{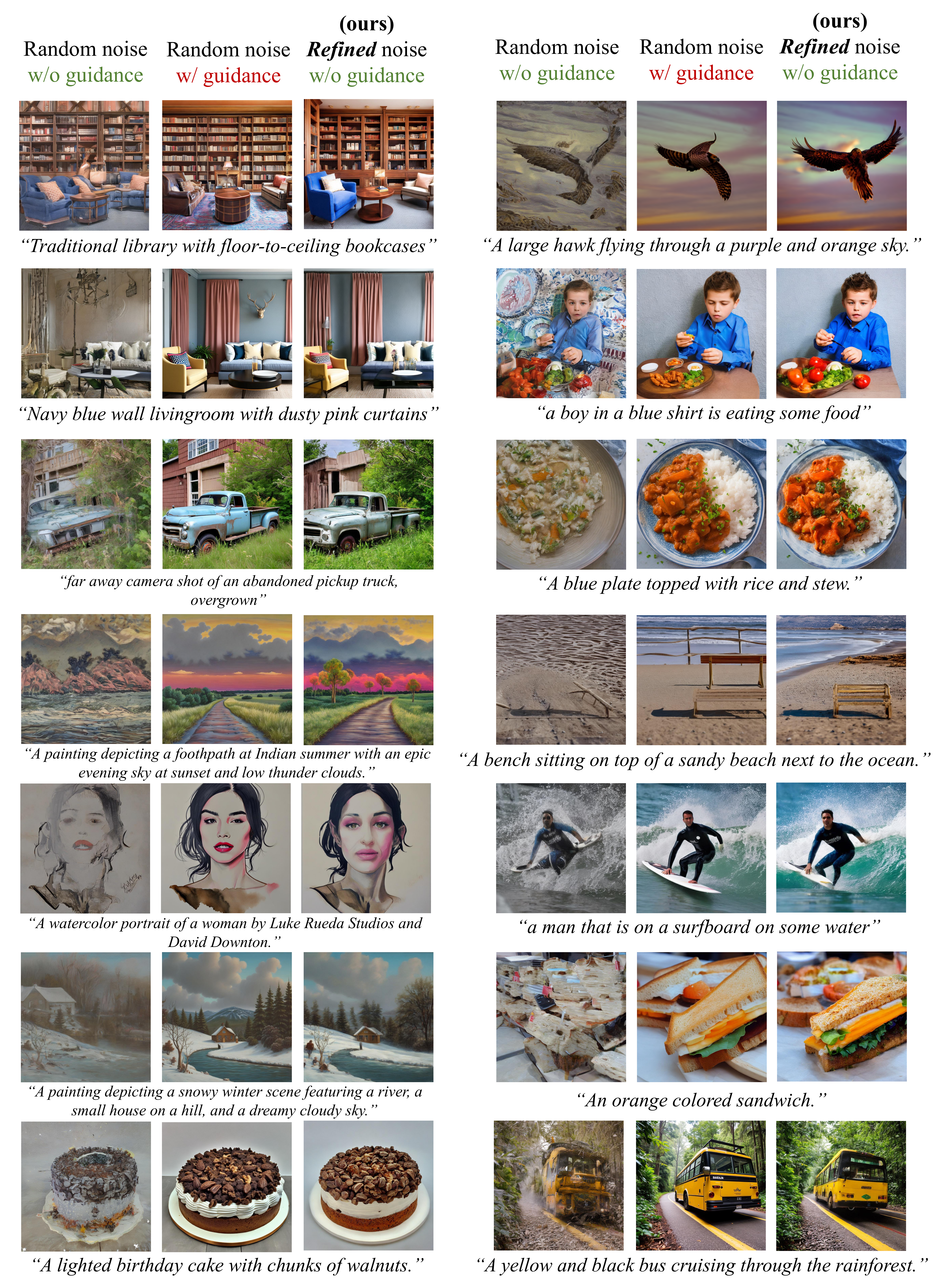}
    \vspace{-10pt}
    \caption{
    \textbf{Additional qualitative results.}
    }
    \label{fig:supp_qual2}
\end{figure}

\begin{figure}[!b]
    \centering
    \includegraphics[height=1.0\textheight]{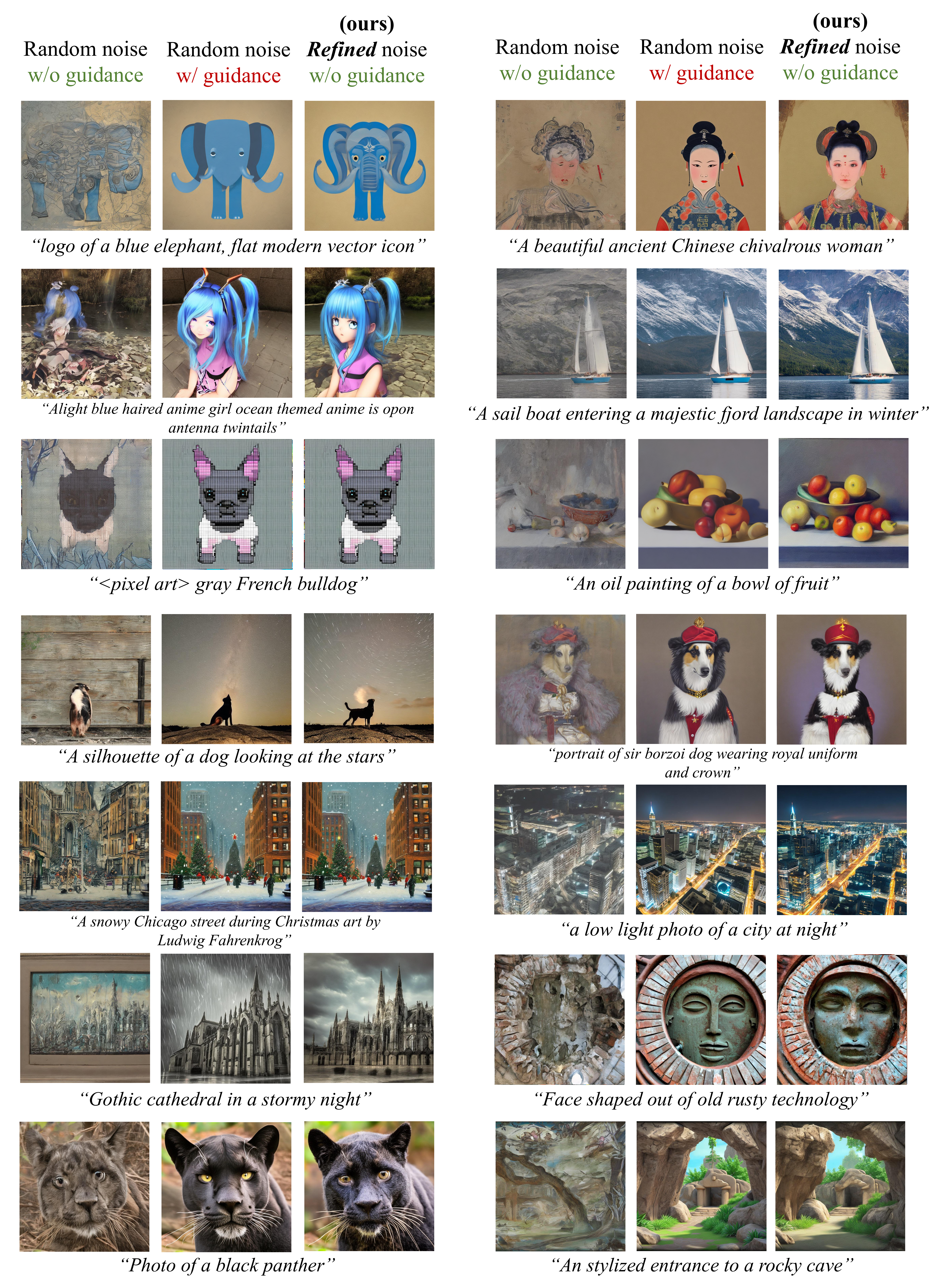}
    \vspace{-10pt}
    \caption{
    \textbf{Additional qualitative results.}
    }
    \label{fig:supp_qual3}
\end{figure}

\begin{figure}[!b]
    \centering
    \includegraphics[height=1.0\textheight]{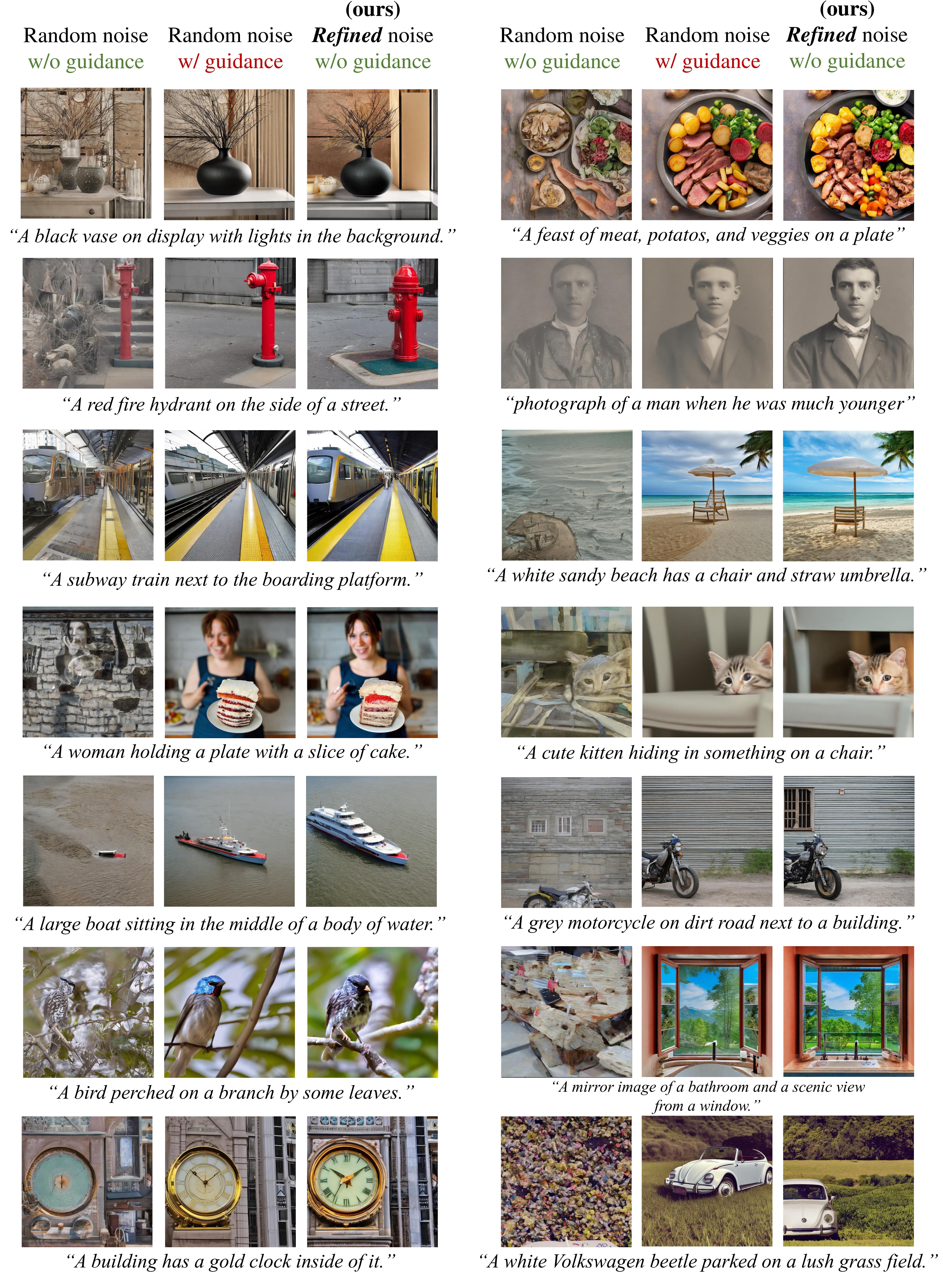}
    \vspace{-10pt}
    \caption{
    \textbf{Additional qualitative results.}
    }
    \label{fig:supp_qual4}
\end{figure}

We present our additional qualitative results on Fig.~\ref{fig:supp_qual},~\ref{fig:supp_qual2},~\ref{fig:supp_qual3} and~\ref{fig:supp_qual4}. Results show that the performance of using refined noise by \ourmodel is comparable to that of using guidance on random Gaussian noise. All the results are selected from images used in Tab.~\ref{table:main_quan} and~\ref{table:main_fid_is}.

\subsection{User Study}
We conducted a user study to evaluate prompt adherence and image quality by comparing images generated from random Gaussian noise and our refined noise. The results are presented in Tab.~\ref{supp:user_scratch}. The study demonstrates that our method outperformed the baseline in all human evaluation criteria. A total of 26 participants anonymously evaluated 20 pairs of images, each pair consisting of an image generated using initial Gaussian noise and our refined noise from \ourmodel. The percentage was calculated by dividing the total number of selections for each option by the total number of responses, following the same methodology as in Tab.~\ref{table:user_study}.

Participants were provided with the following instructions for each pair of images:\\

\begin{enumerate}
    \item Which image has better overall quality? (left/right)
    \item Which image more faithfully reflects the given prompt? (left/right)
\end{enumerate}

\begin{table}[h]
    \centering
    \resizebox{0.6\linewidth}{!}{
    \begin{tabular}{lcc}
        \toprule
        \textbf{Metric} & \textbf{Gaussian Noise} & \textbf{Refined Noise (Ours)} \\
        \midrule
        Image Quality & 3.08\% & \textbf{96.92\%}   \\
        Prompt Adherence &  6.73\% & \textbf{93.27\%}  \\
        \bottomrule 
    \end{tabular}
    }
    \caption{\textbf{User study on the image quality and prompt adherence of generated images.}}
    \label{supp:user_scratch}
    \vspace{-15pt}
\end{table}

\clearpage
\section{Implementation and Experimental Details}
\label{supp:imple_details}

\subsection{Implementation Details}


\label{subsec:details}

\begin{figure}[h]
    \centering
    \includegraphics[width=0.7\linewidth]{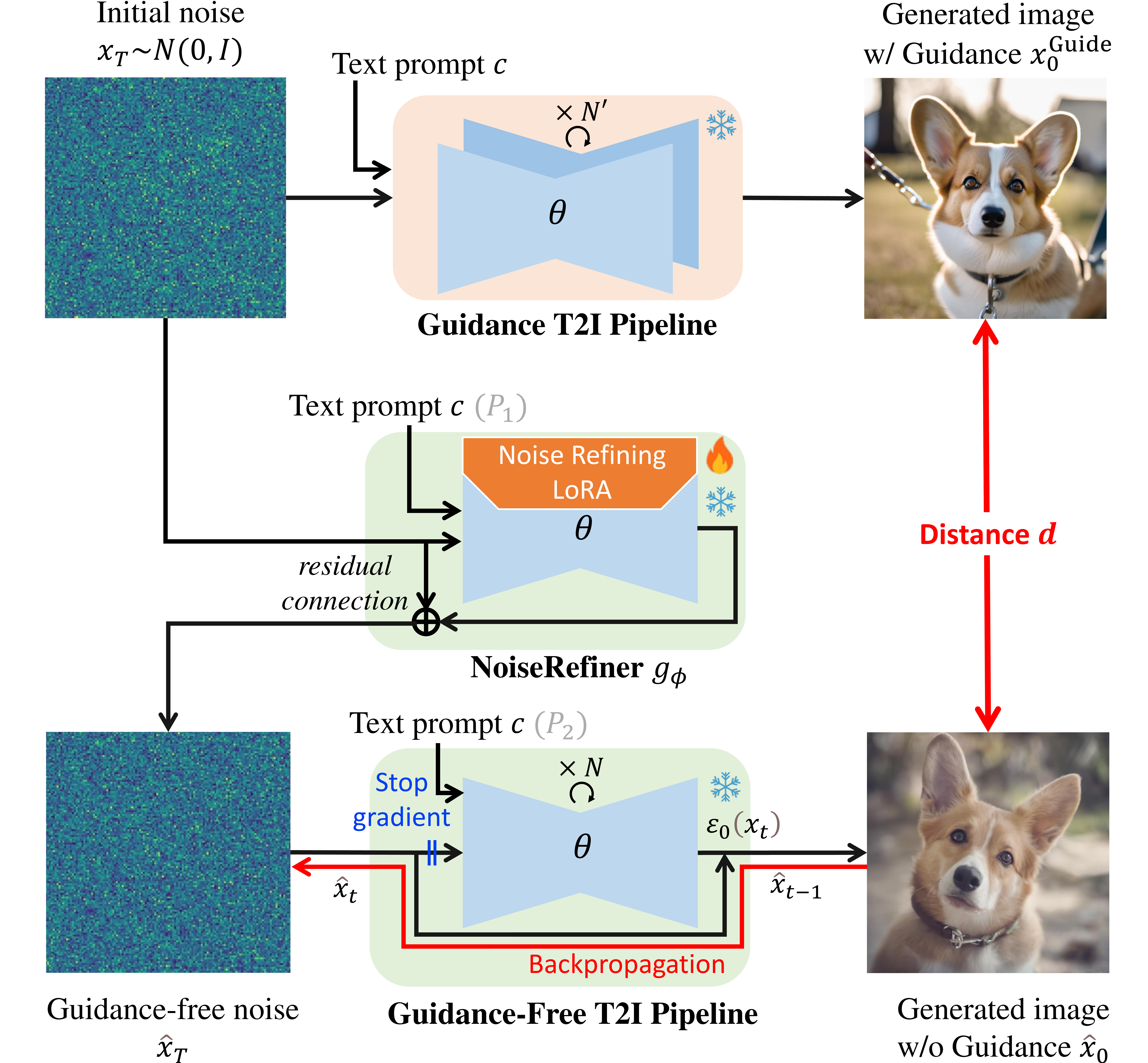}
    \caption{\textbf{Training framework.} We provide an annotated illustration of the training framework to clarify the notation in the following discussion.}
    \label{fig:supp_training_framework}
\end{figure}

\paragraph{More details of our framework.}
We can generalize our framework \textit{NoiseRefine} from pixel-level diffusion models to latent-level diffusion models, but in our experiments, we use MSE loss in latent space for $d(\cfgimage, \hat{x}_0)$. We provide our training framework in Fig.~\ref{fig:supp_training_framework}. It consists of three parts: Guidance T2I Pipeline takes Gaussian noise $x_T\sim\mathcal{N}(0,\mathbf{I})$ and condition (text prompt) $c$ as inputs and generates an image $\cfgimage$ with guidance methods~\citeguidance. \ourmodel $g_\phi$ refines Gaussian noise $x_T$. Guidance-Free T2I Pipeline takes refined noise $\hat{x}_T=g_\phi(x_T)$ and condition (text prompt) $c$ and generates an image $\hat{x}_0$ without guidance. For Guidance T2I Pipeline, with the denoising network $\epsilon_\theta$, we can use the guided score $\epsilon^{\text{CFG}}_{\theta}(x_t, c)$ for CFG~\citep{ho2022classifier} or $\epsilon^{\text{PAG}}_{\theta}(x_t, c)$ for PAG~\citep{ahn2024self} in denoising process as below:
\begin{align}
    \epsilon^{\text{CFG}}_{\theta}(x_t, c) &= {\epsilon}_{\theta}(x_t, c) + w ( {\epsilon}_{\theta}(x_t, c) - {\epsilon}_{\theta}(x_t, \varnothing) ),     \label{eq:cfg} \\
    \epsilon^{\text{PAG}}_{\theta}(x_t, c) &= {\epsilon}_{\theta}(x_t, c) + s ( \epsilon_{\theta}(x_t, c) - \hat{\epsilon}_{\theta}(x_t, c) ), \label{eq:pag} \\
    \epsilon^{\text{CFG,PAG}}_{\theta}(x_t, c) &= {\epsilon}_{\theta}(x_t, c)  + w ( {\epsilon}_{\theta}(x_t, c) - {\epsilon}_{\theta}(x_t, \varnothing) ) + s ( \epsilon_{\theta}(x_t, c) - \hat{\epsilon}_{\theta}(x_t, c) ), \label{eq:cfgpag}
\end{align}
where $w$ and $s$ denote the guidance scale of CFG~\citep{ho2022classifier} and PAG~\cite{ahn2024self}, $c$ denotes the condition, and  $\varnothing$ denotes the null condition (\ie, empty prompt). Note that the perturbed score $\hat{\epsilon}_{\theta}$ is from perturbing the forward process of the denoising network $\epsilon_\theta$~\citep{ahn2024self}. With the denoising step $N'=20$, we can get the guided image $x^{\text{Guide}}_{0}$. Our \ourmodel refines Gaussian noise $x_T$ with $g_\phi$ at timestep $t=T$, which is from the reverse step of DDIM~\citep{song2020denoising} in Eq.~\eqref{eq:ddim_reverse_step}. The output of \ourmodel $g_\phi$ is denoted as  $\hat{x}_T=g_\phi(x_T)$ and becomes the input of Guidance-Free T2I Pipeline. In this pipeline, $\hat{x}_T$ is denoised into $\hat{x}_0$ without guidance using $N$ denoising steps.
\paragraph{Model details.}
 For \ourmodel \( g_\phi \), we use Stable Diffusion 2.1~\citep{rombach2022high} with LoRA~\citep{hu2021lora} rank of 128, applied to all attention, convolutional, and feed-forward layers. We use DDIM~\citep{song2020denoising} scheduler with the same settings as the pre-trained model. For noise refinement, we use an input timestep \( T = 999 \), and the default denoising step \( N \) is set to 10.

\subsection{Experimental Details}

\paragraph{Training details.}
The training dataset consists of two parts: 20K images generated with CFG~\citep{ho2022classifier} (guidance scale 7.5) using prompts from MS COCO~\citep{lin2014microsoft}, and 30K images generated with both CFG~\citep{ahn2024self} (guidance scale 3.0) and PAG~\citep{ahn2024self} (guidance scale 2.0) using prompts from Pick-a-pic~\citep{kirstain2023pick}. Only images scoring above 6.0 on the LAION Aesthetics Predictor V2~\citep{schuhmann2022aestheticpredictor} are selected for training. Furthermore, for training, we use 8 A100 GPUs of 40GB vRAM with a batch size of 4 and sample the images for evaluation using weights of 39k training steps.

\paragrapht{Evaluation details.}
For sampling images with guidance in Tab.~\ref{table:main_fid_is}, 40\% of 30k images are generated with CFG~\citep{ho2022classifier} ($w=7.5$) and remaining 60\% are generated with both CFG~\cite{ho2022classifier} (uniformly from $w\in [3.0, 5.0]$) and PAG~\citep{ahn2024self} (uniformly from $s\in [2.0, 3.0]$). For Tab.~\ref{table:main_quan}, we use 200 prompts from Drawbench~\citep{saharia2022photorealistic}, 400 prompts from HPDv2~\citep{wu2023human}, and 500 prompts from test set of Pick-a-pic~\citep{kirstain2023pick} for generating 5 images per prompt. Here, for MS-COCO~\citep{lin2014microsoft}, we use 5k generated images selected from those used in Tab.~\ref{table:main_fid_is}.  The qualitative results for Fig.~\ref{fig:main_qual} are from the images used in Tab.~\ref{table:main_quan} and~\ref{table:main_fid_is}. Additionally, \textit{Inference time} in Tab~\ref{table:main_fid_is} is computed by averaging time per image across 30K images generated with the inference step of 20 and a batch size of 1 on RTX 3090.

\paragrapht{Ablation study settings.}
For the ablation study on the number of denoising steps, we use the training dataset which consists of MS COCO~\citep{lin2014microsoft} and Pick-a-pic~\citep{kirstain2023pick} used in training \ourmodel. The models are trained in two V100 GPUs for 100K steps. In the case of training dataset filtering, we only use MS-COCO~\citep{lin2014microsoft} dataset, and the models are trained in two RTX 3090 GPUs for 25K steps. For the unfiltered case, entire 80K images are used. For the filtered case, we use the same filtering criteria detailed in~\ref{subsec:details}, resulting in 20K images. All the other training details are kept consistent.



\clearpage
\section{Discussion}
\label{supp:discussion}
In this section, we compare the performance between training the \ourmodel $g_\phi$ and the denoising network $\epsilon_\theta$ in the denoising process without guidance (Sec.~\ref{sec:supp:whyprompt}). In addition, we present our hypothesis on why refined noise eliminates the need for guidance methods, explaining it step by step (Sec.~\ref{sec:supp:whyrefined}). We further analyze the impact of initial noise and prompt on the generated image (Sec.~\ref{sec:supp:impactofnoise}). 

\subsection{Effectiveness of Prompt Learning}
\label{sec:supp:whyprompt}
As shown in the training framework of our method (Fig.~\ref{fig:supp_training_framework}), the \ourmodel can be trained using the loss \( d(\cfgimage, \hat{x}_0) \), but the denoising network \( \epsilon_\theta \) within the Guidance-Free T2I pipeline can also be trained. Instead of directly training the model itself (e.g., fine-tuning models like CLIP~\citep{radford2021learning}), our method demonstrates the efficiency of learning the noise input to the Guidance-Free T2I pipeline, akin to \textit{prompt learning}, which optimizes prompts instead of models. Specifically, similar to \textit{conditional prompt learning} such as CoCoOp~\citep{zhou2022conditional}, noise prompts are conditionally generated based on different inputs (Gaussian noise \(x_T\) and text prompt \(c\)).

By leveraging the knowledge of pretrained denoising networks, noise prompts can be generated efficiently, as verified in Sec.~\ref{sec:supp:pretrained}. Here, we examine the case of training Guidance-Free T2I pipeline itself. Specifically, following the settings of ablation study on the number of denoising steps (Tab.~\ref{table:ablation_table}) where we use filtered MS COCO~\citep{lin2014microsoft} dataset, we compare the performance of models trained using our learning method (Fig.~\ref{fig:supp_training_framework}) with models where \( \epsilon_\theta \) is fine-tuned on \(x_T\) as input without using the \ourmodel. Instead of directly fine-tuning \( \epsilon_\theta \), we train a LoRA~\citep{hu2021lora} module with the same rank and layers as \( g_\phi \) (used in the \ourmodel).

The results in Fig.~\ref{fig:model-tuning} clearly show that training the noise prompt leads to significantly faster convergence and higher-quality outputs at the same training steps. In the figure, the first row represents the outputs from models trained with the \ourmodel $\epsilon_\theta$, while the second row shows outputs from models where \( \epsilon_\theta \) was fine-tuned using LoRA~\citep{hu2021lora}.

\subsection{Why does refined noise help denoising?}
\label{sec:supp:whyrefined}

To identify which refined noise components contribute to guidance-free generation, we first decompose the refined noise into multiple frequency components. In this study, we utilize a two-dimensional Fourier transform to break down both the refined noise and the initial noise into their respective frequency components. Each frequency component is represented by a frequency band, denoted as $(a,b)$, which corresponds to the frequency range from $a$ to $b$. Note that although we explored other decomposition methods, such as dividing the noise into patches, they did not yield interpretable results.

\begin{figure}[h]
    \centering
    \includegraphics[width=1.0\linewidth]{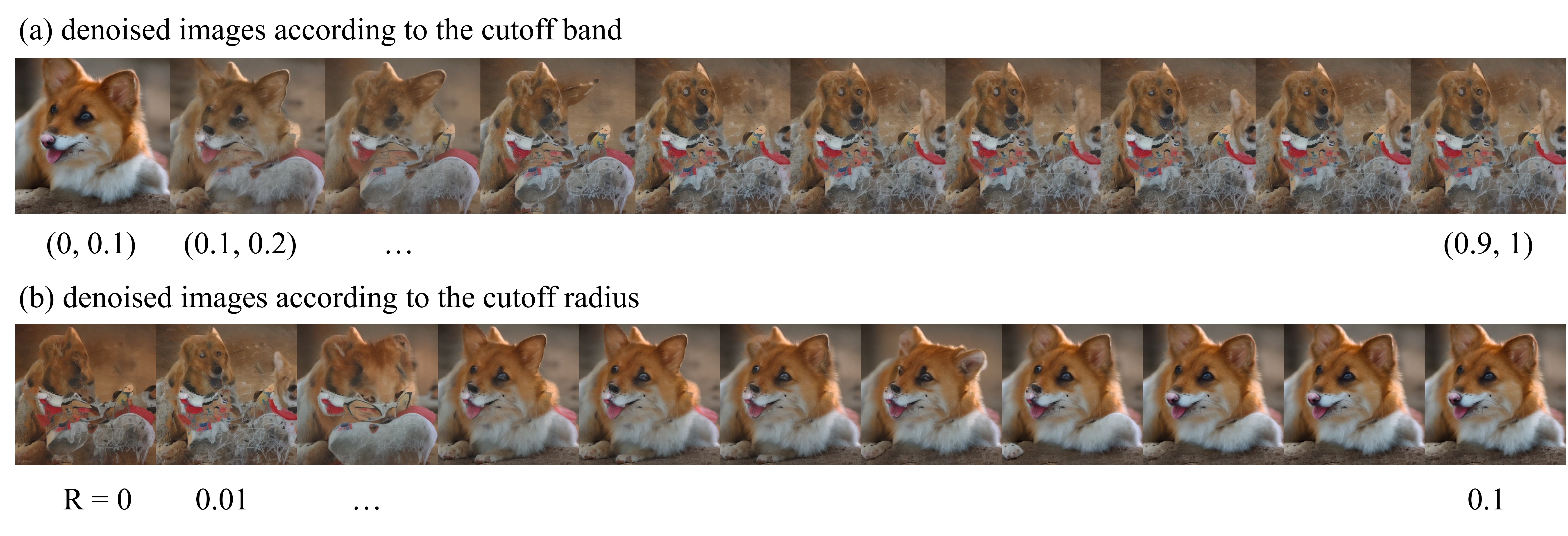}
    \vspace{-10pt}
    \caption{\textbf{Visualization of denoised images according to the cutoff band.} Both refined and initial noise were transformed into the frequency domain using Fourier transforms. The frequency domain of the initial noise, normalized such that the maximum radius is 1. (a) The frequency divided into intervals of 0.1. For each interval, the corresponding frequency components were replaced with those from the refined noise, followed by denoising. The results show that only when the (0, 0.1) frequency band was replaced does an image generated by the refined noise emerge. (b) Visualization of denoised images by incrementally increasing the cutoff radius $R$ from 0 in steps of 0.01 and replacing the corresponding components of the initial noise with refined noise. The results demonstrate that images denoised using refined noise are obtained starting at a cutoff radius of 0.03.}
    \label{fig:fft_band_image}
    \vspace{-10pt}
\end{figure}

\begin{figure}[h]
    \centering
    \includegraphics[width=0.5\linewidth]{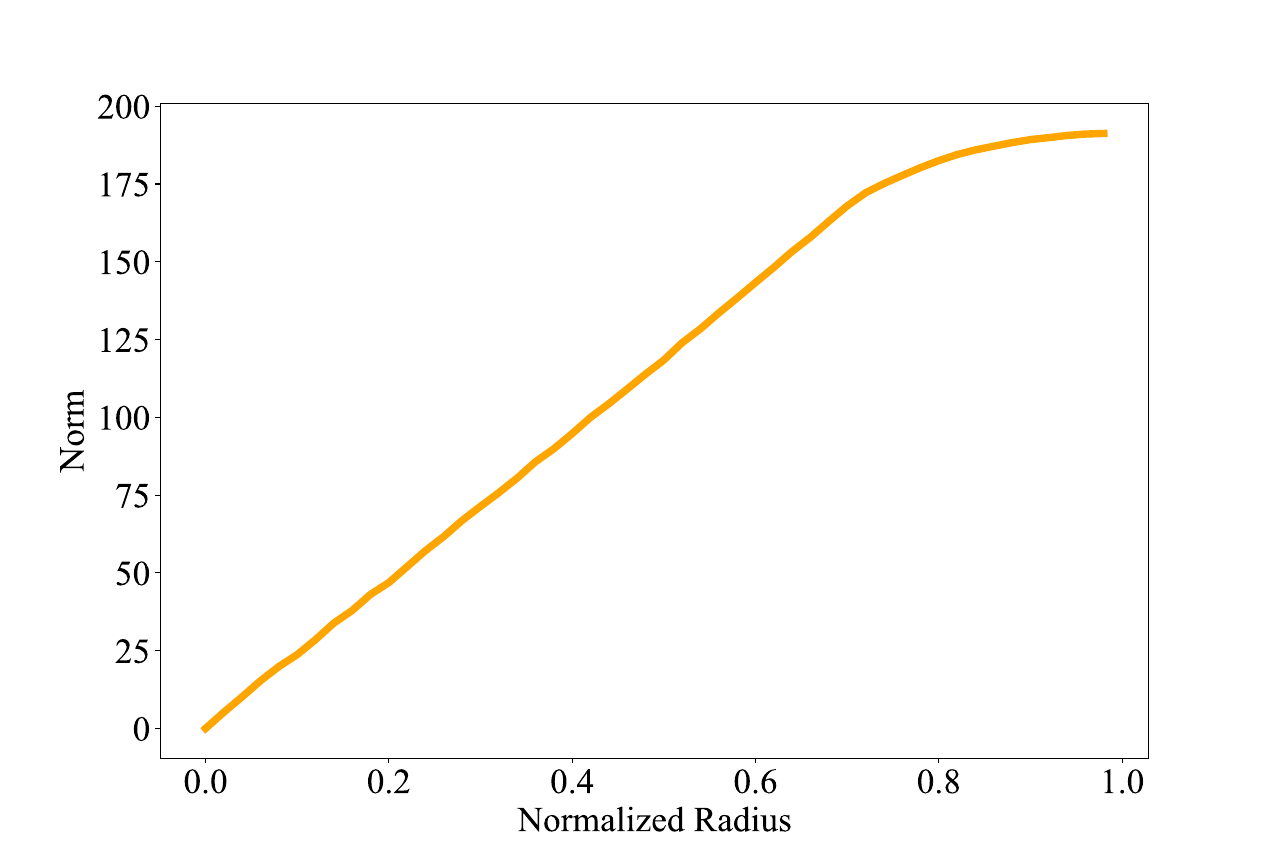}
    \caption{\textbf{Visualization of the norm based on the frequency-filtered radius $R$ of refined noise.} This visualization demonstrates the increase in the norm as the cutoff radius $R$ in the frequency domain is expanded. The refined noise was transformed into the frequency domain using a Fourier transform, and the norm corresponding to each cutoff radius was calculated and plotted. }
    \label{fig:norm_graph}
\end{figure}

\paragraph{Low-frequency components matter.}
Using 2D Fourier transforms, we transform both refined and initial noise into the frequency domain. The initial and refined noise frequency domain is normalized into $(0,1)$. We synthesize a new noise signal by replacing specific frequency bands of the initial noise with the corresponding bands from the refined noise. Fig.~\ref{fig:fft_band_image} (a)
 presents the generated images corresponding to different frequency bands, demonstrating that the low-frequency components of the refined noise predominantly influence the generation process. In Fig.~\ref{fig:fft_band_image} (b), images are generated by varying the band length within the low-frequency region. The results indicate that, despite the low magnitude of the low-frequency components, which can be confirmed through Fig.~\ref{fig:norm_graph}, they are sufficient to reconstruct the image effectively.

\begin{figure}[h]
    \centering
    \includegraphics[width=1.0\linewidth]{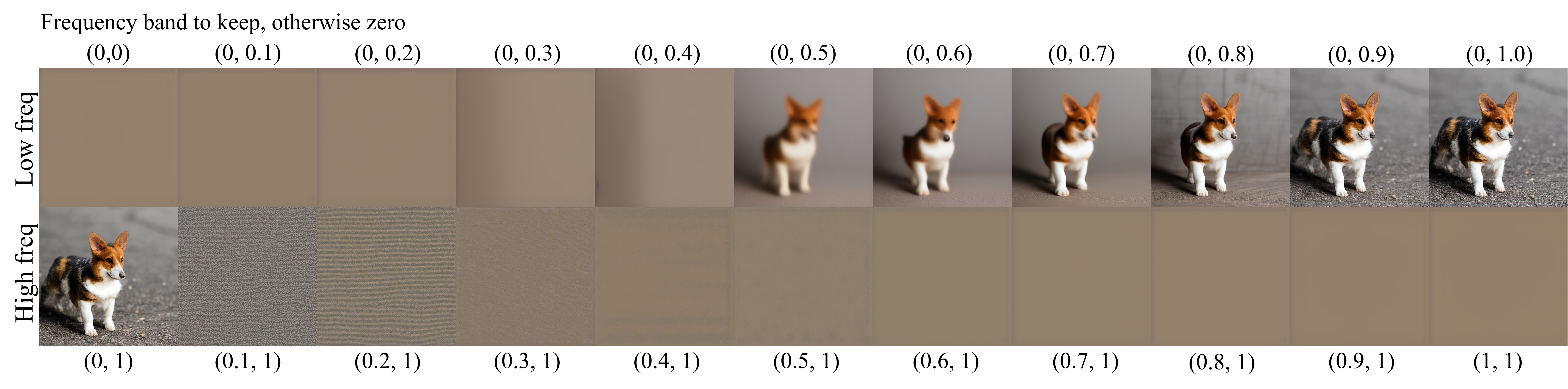}
    \caption{\textbf{Denoised images using only low(top) / high(bottom) frequency components.} Diffusion models can generate the overall structure of the image using only the low-frequency bands of the refined noise. We use DDIM~\citep{song2020denoising} with 20 steps for denoising without CFG, and the prompt was \textit{``a photo of a corgi''}.}
    \label{fig:supp_low_high_freq}
\end{figure}

\paragrapht{Diffusion models can generate images using only low-frequency components.}  
In Fig.~\ref{fig:supp_low_high_freq}, we examine how well diffusion models can denoise when specific frequency bands of refined noise are retained, and the values of the remaining bands are set to zero (using ideal high/low pass filters). The top row shows the results of applying a 2D Fourier transform to the refined noise, normalizing the FFT frequency domain into $(0,1)$, and sequentially retaining lower frequency bands, such as $(0, 0), (0, 0.1), (0, 0.2), ..., (0, 1)$, while setting the remaining bands to zero. These noise inputs are then denoised without CFG~\cite{ho2022classifier}. The figure demonstrates that the diffusion model begins forming a recognizable corgi shape even when only the lower 50\% of frequency bands of the refined noise are present. In contrast, noise containing only high-frequency bands fails to generate coherent images.

\vspace{10pt}

\begin{figure}[h]
    \centering
    \includegraphics[width=1.0\linewidth]{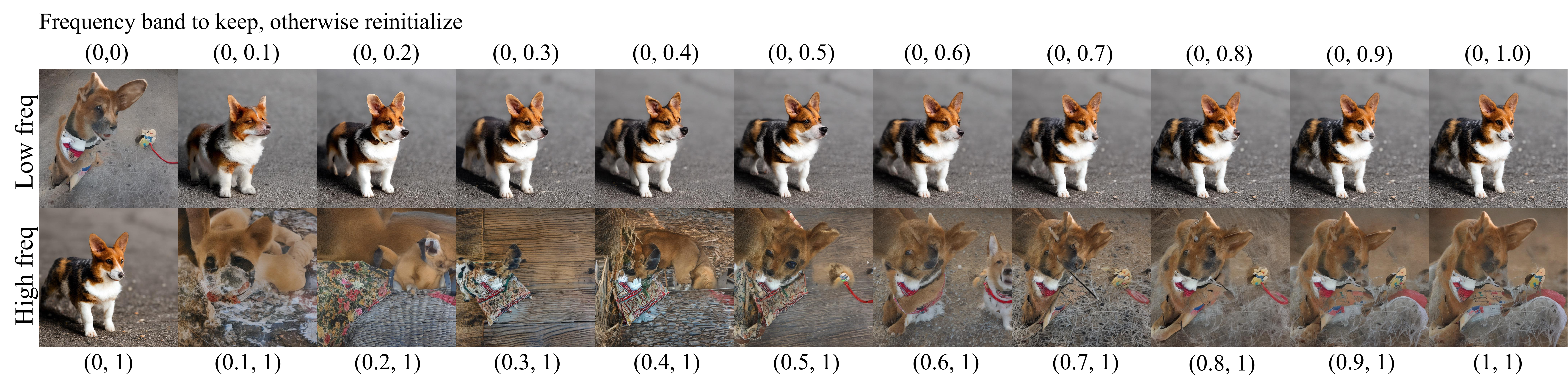}
    \caption{\textbf{Denoised images using only low (top) / high (bottom) frequency components with reinitialization.} We use DDIM~\citep{song2020denoising} with 20 steps for denoising without CFG, and the prompt was \textit{``a photo of a corgi''}.}
    \label{fig:supp_low_high_freq_reinit}
    \vspace{-20pt}
\end{figure}

\paragraph{High-frequency components contribute details.}  
Here, we use the same noise decomposition process of refined noise as Fig.~\ref{fig:supp_low_high_freq} but following~\citep{geng2025factorized}, we reinitialize the frequency components that were set to zero with corresponding components from standard Gaussian noise, then denoise again. The results, shown in Fig.~\ref{fig:supp_low_high_freq_reinit}, indicate that when all frequency components are present, the diffusion model can generate clear and complete images. Randomly reinitialized high-frequency components appear to add details onto the structure formed by the low-frequency components. While refined noise retaining only the lower 10\%–20\% of frequencies can still reconstruct the original image when the rest is reinitialized, noise retaining only the high-frequency components fails to do so. This suggests that low-frequency components alone carry the significant information needed for image generation.

\begin{figure}[h]
    \centering
    \begin{subfigure}[b]{0.49\textwidth}
        \centering
        \includegraphics[width=\textwidth]{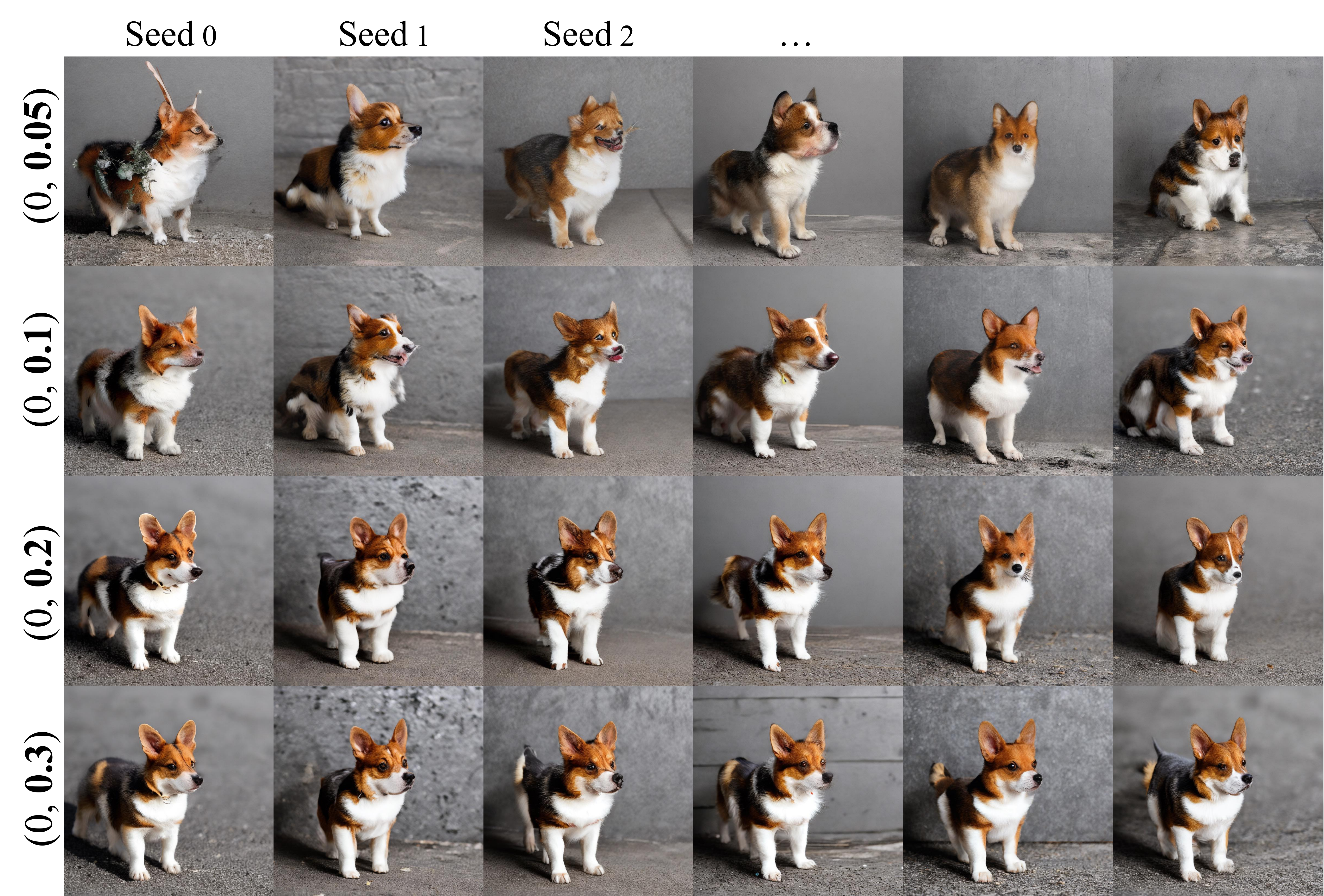}
        \caption{Low-frequency components of refined noise}
        \label{fig:subfig1}
    \end{subfigure}
    \hfill 
    \begin{subfigure}[b]{0.49\textwidth}
        \centering
        \includegraphics[width=\textwidth]{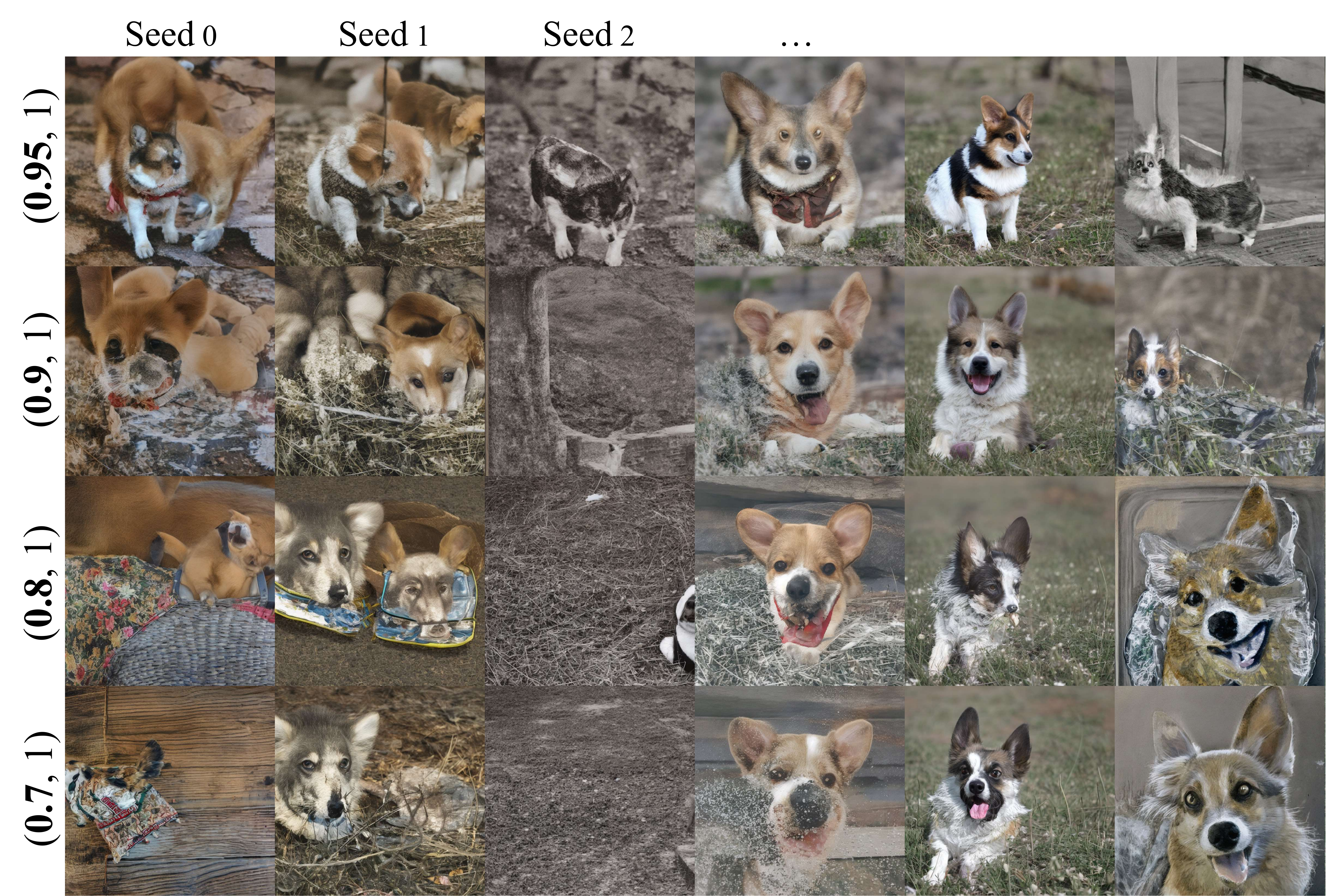}
        \caption{High-frequency components of refined noise}
        \label{fig:subfig2}
    \end{subfigure}
    \caption{\textbf{Different denoised images using only low(a) / high(b) frequency components for different seeds.} Here we use 8 different seeds. From the top rows, it visualizes 8 images using only the lower (a) / higher (b) 5\%, 10\%, 20\%, and 30\% (from the top to the last rows) frequency components of the refined noise.}
    \label{fig:diff-seeds}
    \vspace{-10pt}
\end{figure}

In Fig.~\ref{fig:diff-seeds}, each row visualizes images generated with only the lower 5\%, 10\%, 20\%, and 30\% (from the top rows to last rows) frequency components of the refined noise, while the bottom row shows images generated with only the upper 5\%, 10\%, 20\%, and 30\% frequency components. These results confirm that low-frequency components encode the overall layout and structure, whereas high-frequency components lack meaningful information.

From these observations, we infer that the poor quality of unguided diffusion model outputs is due to their failure to form appropriate low-frequency components during denoising. High-frequency details added on poorly formed layouts result in artifacts that are perceived as unnatural.

\paragrapht{How do guidance methods form plausible initial layouts?} 
As highlighted in \citep{ahn2024self}, classifier-free guidance (CFG)~\citep{ho2022classifier} enhances the difference between conditional and unconditional predictions at each step, amplifying ``signals that can only be generated with the condition'' (\eg, features like the eyes or nose of a corgi in \textit{``a photo of a corgi''}). This effectively strengthens salient features corresponding to low-frequency components in the early denoising steps. From this, we deduce that guidance methods~\citep{ahn2024self,ho2022classifier,hong2023improving} add appropriate low-frequency components during inference, aiding the formation of high-quality layouts.

\begin{figure}[h]
    \centering
    \includegraphics[width=1.0\linewidth]{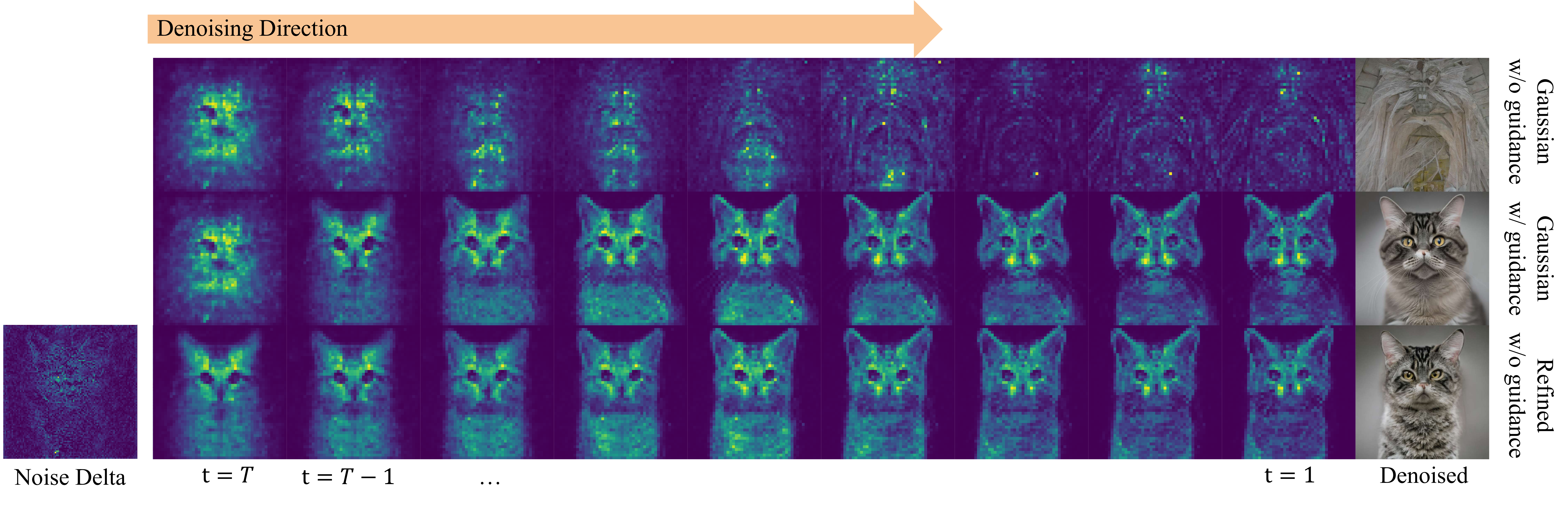}
    \vspace{-20pt}
    \caption{
    \textbf{Visualization of 11th layer cross attention map.}
    Token corresponding to `cat' is used for visualization among the prompt `a photo of a cat'. First and second row is the case starting from random Gaussian noise where guidance is not used for the first row and used for the second row. Third row is the starting from refined noise by the \ourmodel. `Noise Delta' means the difference between initial Gaussian noise $x_T$ and refined noise $\prednoise$. When guidance is not used, failure to create meaningful attention map across all timestep is notable, leading to completely broken generation. However when guidance or our refined noise is used, meaningful cross attention map is observed, leading to successful generation. Notably thanks to noise delta, a better aligned cross attention map is observed even in earlier step ($t=T$) when the refined noise is used.
    }
    \label{fig:crossattn}
    \vspace{-10pt}
\end{figure}

\paragrapht{How does the \ourmodel form low-frequency layouts?}  
Interestingly, the \ourmodel naturally forms low-frequency layouts even though our training framework does not explicitly enforce learning them as can be seen in Fig.~\ref{fig:analysis_pred}. To understand this, we analyze cross-attention and self-attention maps across denoising steps. Fig.~\ref{fig:crossattn} visualizes these maps at different timesteps. Gaussian noise fails to form meaningful cross-attention maps in early steps due to its near-zero signal-to-noise ratio (SNR), which is expected. However, this failure persists in later steps, indicating an inability to form well-aligned layouts (Fig.~\ref{fig:crossattn} first row).

Research \citep{chefer2023attend,guo2024initno,mao2023semantic} has shown that reducing noisy artifacts in cross-attention maps and aligning them with object regions during inference improves performance. This suggests that the failure of cross-attention maps to align is a key reason for the diffusion model’s inability to create coherent layouts. When using CFG~\citep{ho2022classifier} (second row) or refined noise (third row), the cross-attention maps align well with the prompt, resulting in better outputs. Notably, cross-attention maps for refined noise exhibit accurate object shapes from the very first step, implying that the diffusion model can form plausible layouts from the beginning of the denoising process. This is further supported by Fig.~\ref{fig:noise_denoising_process}, which visualizes \(x_0\) predictions at each denoising step.

\paragraph{Implications for guidance-free generation.}  
Without guidance methods or noise refiners aiding the formation of low-frequency layouts, diffusion models fail to create plausible initial layouts. Random low-frequency components lead to artifacts that are perceived as unnatural. An interesting avenue for future research would be identifying why diffusion models struggle to form low-frequency components without guidance and developing training techniques to eliminate the need for guidance during the training stage.

\clearpage
\subsection{Impact of Initial Noise and Prompt on Generated Image}
\label{sec:supp:impactofnoise}
\begin{figure}[h]
    \centering
    \includegraphics[width=0.8\linewidth]{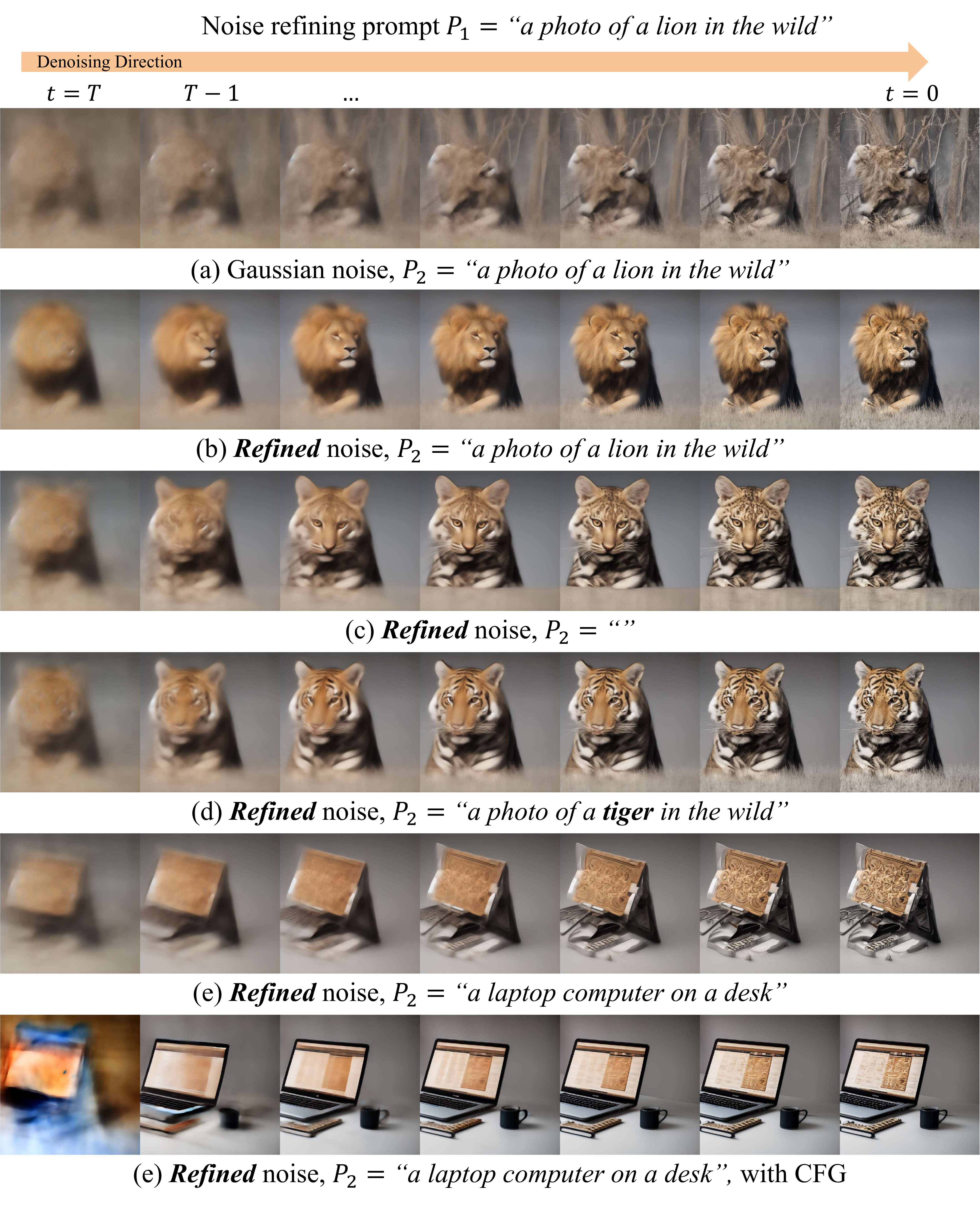}
    \caption{\textbf{Visualization of denoised image using different prompt for noise refinement $\epsilon_\theta$ and denoising $g_\phi$.}}
    \label{fig:diff_prompt}
\end{figure}

We previously demonstrated how refined noise affects initial layouts and how guidance and refined noise contribute to forming these layouts effectively. In this section, we investigate how the `layout' and the prompt influence the final generated image during the denoising process. 
Specifically, we explore what happens when the prompt used to generate the initial layout (\(P_1\), one of the inputs to the \ourmodel \(g_\phi\)) differs from the prompt used during denoising (\(P_2\), one of the inputs to the denoising network \(\epsilon_\theta\) in the Guidance-Free T2I Pipeline shown in Fig.~\ref{fig:supp_training_framework}). 
Does the model prioritize one prompt over the other? Or does it attempt to harmonize both? We investigate this question through the results shown in Fig.~\ref{fig:diff_prompt}.

\begin{itemize}
    \item Fig.~\ref{fig:diff_prompt} (a) visualizes the predicted $x_0$ term in Eq.~\ref{eq:ddim_reverse_step} during the denoising process when no layout is provided (starting from Gaussian noise). The leftmost image corresponds to the predicted $x_0$ at $t=T$, and subsequent images are visualized every three steps. Due to the noisy and ambiguous nature of the initial layout of Gaussian noise, the diffusion model fails to form a coherent lion layout from the initial structure. Instead, it partially adds features such as fur, mane, nose, or mouth, resulting in poor perceptual quality. 

    \item In contrast, (b) shows that in the case of $P_1 = P_2$, refined noise effectively forms the lion layout from the beginning. The diffusion model accurately places the overall lion shape, including its mane, eyes, nose, and mouth, in appropriate positions during the denoising process.

    \item (c) shows the results when the denoising prompt $P_2$ is set to an empty prompt (null prompt). Despite this, the model successfully generates a feline animal based solely on unconditional generation, as the layout sufficiently captures the overall structure of the object. This can be interpreted as the information embedded in the \textbf{\textit{refined}} noise. 

    \item (d) demonstrates the case where the denoising prompt $P_2$ is set to a prompt similar to the initial layout prompt (\textit{``a photo of a tiger in the wild''}). When a similar prompt is used, the image retains the layout provided by the refined noise while also adhering to the prompt. 

    \item In (e), $P_2$ is set to an entirely independent prompt (\textit{``a laptop computer on a desk''}). Here, the model fails to generate a coherent image corresponding to the layout or the prompt. The diffusion model attempts to form a laptop on the existing lion or feline layout but fails to align with the laptop prompt, leading to failure.

    \item Finally, (f) shows that applying CFG~\citep{ho2022classifier} in the settings of (e) allows the diffusion model to disregard the initial layout and generate a laptop. This partially explains why CFG consistently produces high-quality images. Randomly generated initial noise is unlikely to align with the prompt (as shown in (a)), and CFG helps the model ignore such initial noise and generate images consistent with the given prompt.
\end{itemize}

\paragraph{Interpolation between refined noise.}

\begin{figure}[h]
    \centering
    \vspace{-20pt}
    \includegraphics[width=0.7\linewidth]{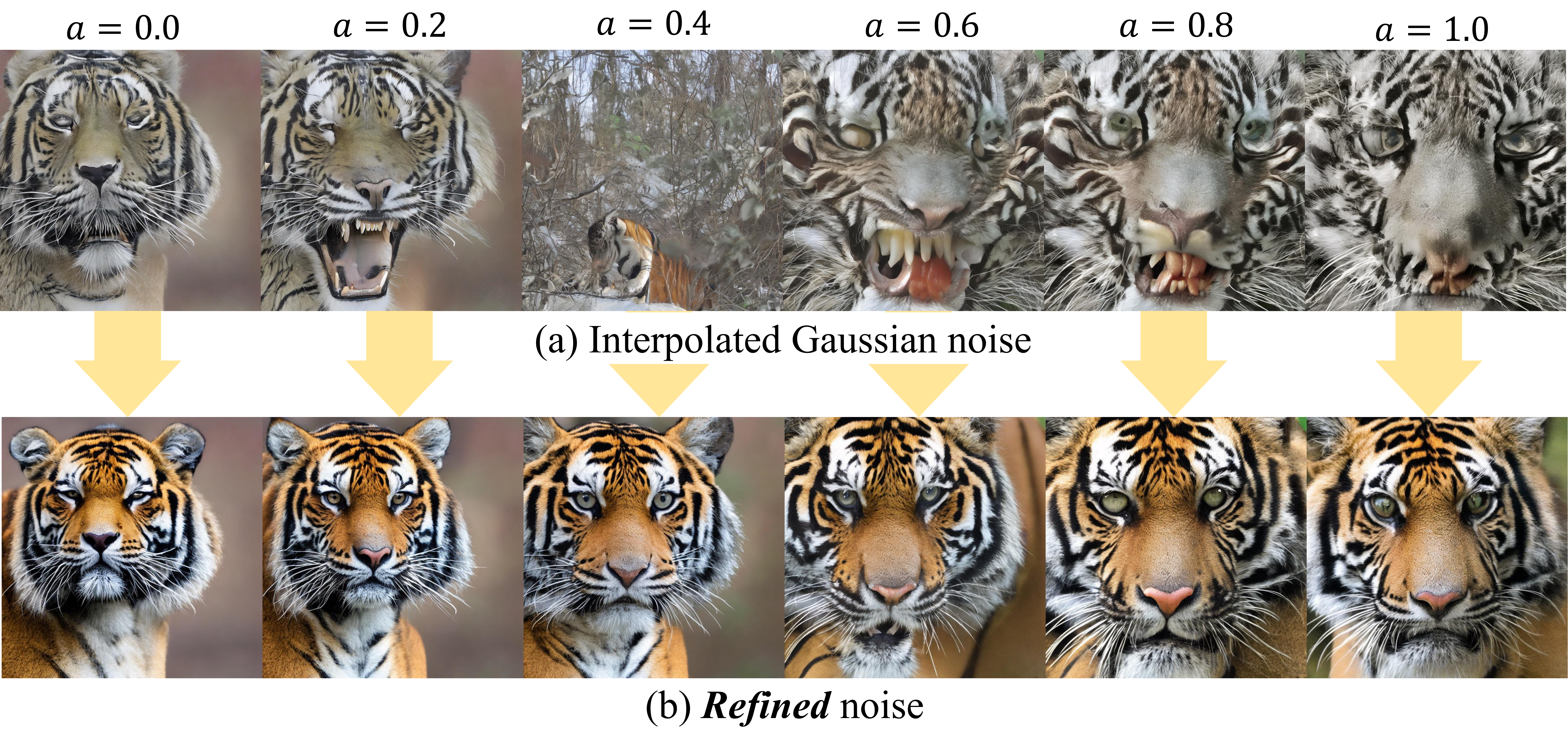}
    \caption{\textbf{Images from interpolated refined Gaussian noise.}}
    \label{fig:interpolation}
\end{figure}

To evaluate whether the \ourmodel effectively learns noise mapping, we follow \citep{song2020denoising, song2020score} to perform spherical interpolation on initial noise samples, generating multiple interpolated noises. We then refine each interpolated noise using the \ourmodel and verify that the refined noises effectively interpolate natural images. In Fig.~\ref{fig:interpolation}, (a) shows the images denoised by the diffusion model without any guidance method, starting from spherical interpolations of two random Gaussian noises. Specifically, each interpolated noise is obtained by performing \text{slerp}($x_{T_1}$, $x_{T_2}$, $a$) for various interpolation ratios $a$, where \text{slerp} performs spherical interpolation between two Gaussian noise at a ratio of $a$.

Fig.~\ref{fig:interpolation} (b) shows the results of denoising the refined versions of these interpolated noises without guidance. The results demonstrate that the refined noises effectively interpolate between the two images. This indicates that the \ourmodel does not simply memorize specific low-frequency signals while ignoring the input noise. Instead, it effectively learns a mapping from a Gaussian noise space to a guidance-free noise space where semantic interpolation between guidance-free images is possible.

\subsection{Comparison with a related work}

A recent study~\citep{kim2024model} exists under the category of noise manipulation. To the best of our knowledge, this work is unique in its focus on learning the noise space itself, rather than optimizing or selecting. Therefore, we compare our proposed approach with this methodology PAHI (Prompt Adaptive Human preference Inversion)~\citep{kim2024model} in this section. 

There are several key differences between the two approaches. First, the tasks being addressed are distinct. While PAHI~\citep{kim2024model} aims at generating outputs aligned with human preferences, our objective is to replace conventional guidance mechanisms entirely. Second, our method offers much greater flexibility. PAHI~\citep{kim2024model} assumes that sampling from certain \(\mathcal{N}(\mu, \Sigma)\) instead of a standard normal Gaussian distribution is more beneficial and predict $\mu$ and $\Sigma$. However, this assumption lacks a strong theoretical foundation. In contrast, our approach aims to learn a gaussian-free noise space without imposing such constraints. Additionally, while PAHI~\citep{kim2024model} is limited to few-step models due to the computational overhead of backpropagation, our approach leverages MSD loss, enabling the use of full-step models without modification. 

\begin{figure}[H]
    \centering
    \begin{minipage}{0.5\textwidth} 
        \centering
        \includegraphics[width=1.0\linewidth]{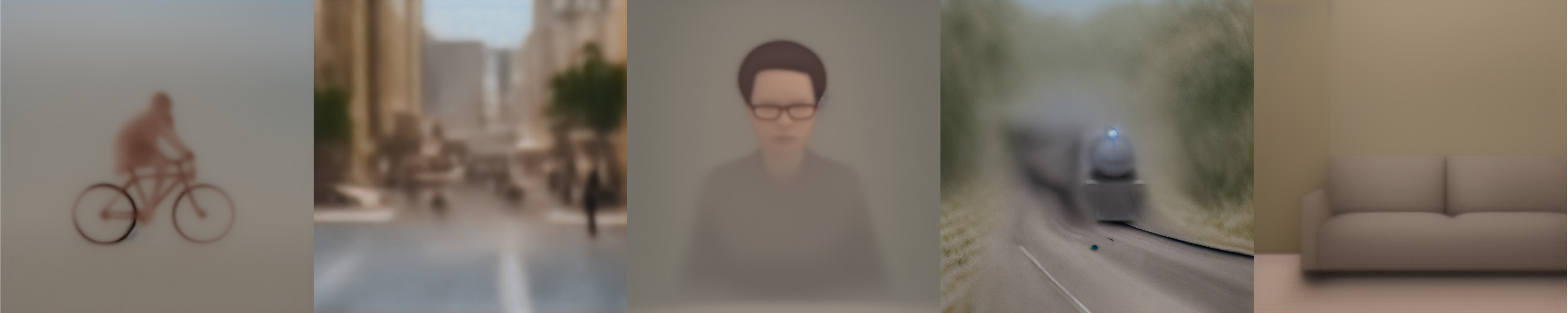}
        \caption{
        \textbf{Qualitative results when employing a shallow 2-layer MLP for estimating Gaussian parameters, as proposed by~\citep{kim2024model}.} 
        The results are significantly blurry, indicating that the simple approach of predicting \(\mu\) and \(\Sigma\) under the assumption that the optimal noise lies within \(\mathcal{N}(\mu, \Sigma)\) performs poorly.
        }
        \label{supp:qual_relwork}
    \end{minipage}
    \hfill 
    \begin{minipage}{0.45\textwidth} 
    \centering
    \begin{tabular}{ l | c }
    \toprule
        \textbf{Method} & \textbf{FID} \\
    \midrule
        MLP~\cite{kim2024model} estimating Gaussian parameters & 217.30 \\
        Noise refining model (ours) &  13.74 \\
    \bottomrule
    \end{tabular}
    \captionof{table}{
    \textbf{Quantitative results when employing a shallow 2-layer MLP for estimating Gaussian parameters, as proposed by~\citep{kim2024model}.}
    }
    \label{supp:quan_relwork}
    \end{minipage}
\end{figure}

Although the official code for PAHI~\citep{kim2024model} is unavailable, we adhere to the guidelines presented in their paper as possible and compare with our method. Specifically, we compare the \ourmodel with the setup that samples noise from \(\mathcal{N}(\mu, \Sigma)\) where $\mu$ and $\Sigma$ is predicted by MLP for a given prompt. Both models are trained with filtered 20K MS COCO\citep{lin2014microsoft} dataset for 25K steps using two RTX 3090 GPUs. Example qualitative results of employing MLP are presented in Fig.~\ref{supp:qual_relwork}, and quantitative comparisons are shown in Tab.\ref{supp:quan_relwork}. Across both evaluations, the \ourmodel outperforms the other setup by a significant margin, showing the effectiveness of our proposed method.

\subsection{Robustness to the number of denoising steps and schedulers}

\begin{figure}[!h]
    \centering
   \includegraphics[width=0.85\textwidth]{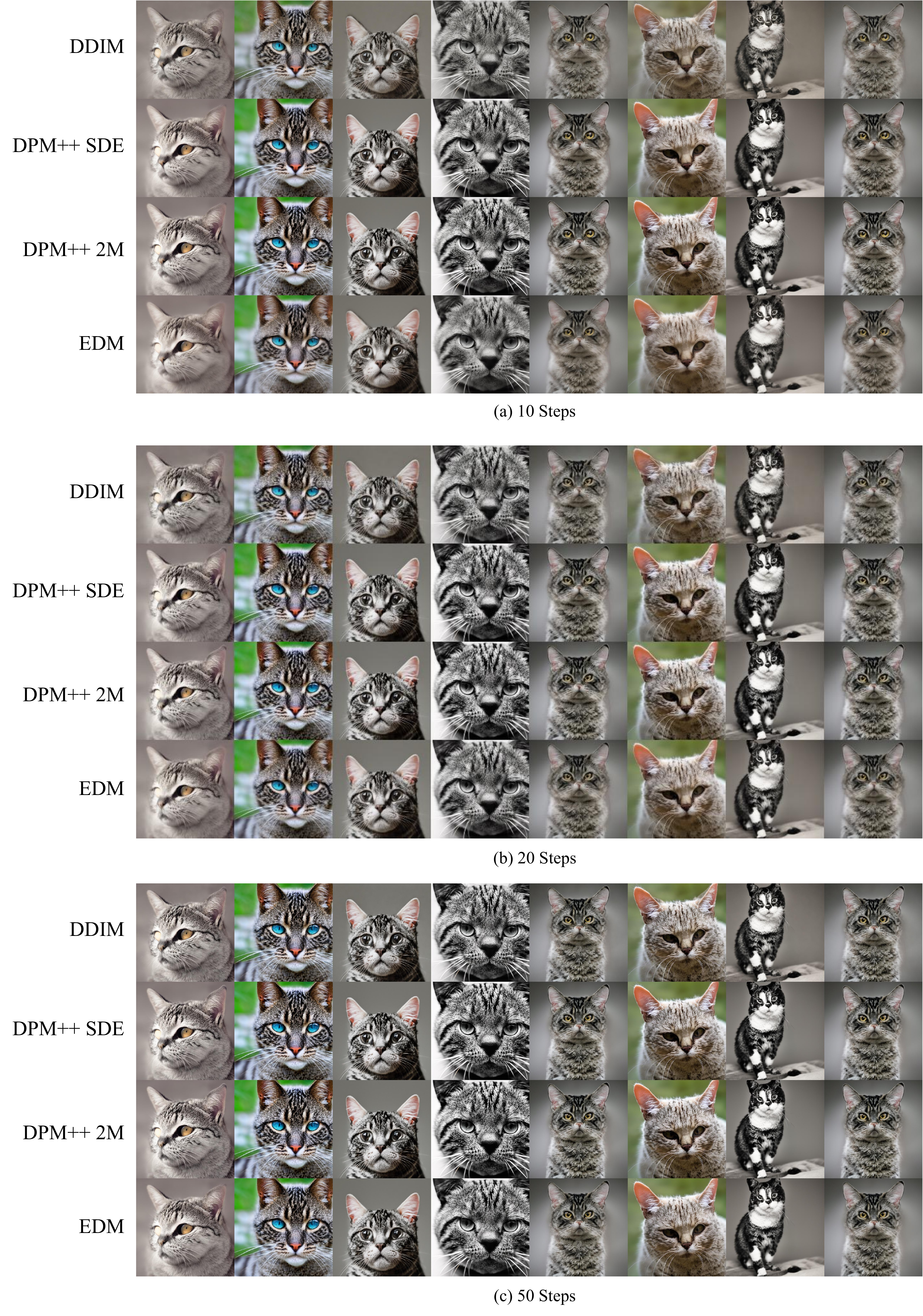}
    \caption{
    \textbf{Inference results on our refined noise in various denoising steps and scheduler settings.}
    (a), (b), and (c) present inference results employing different schedulers at denoising steps of 10, 20, and 50, respectively. The consistency observed across these results highlights the robustness of our refined noise to variations in both denoising steps and schedulers.
    }
    \label{fig:supp_scheduler}
\end{figure}

Since the \ourmodel is trained with a fixed scheduler (DDIM~\citep{song2020denoising}) and denoising steps (10), concerns arise regarding its performance when using different schedulers or denoising steps. To examine the impact of varying schedulers and denoising steps, we conduct experiments comparing qualitative results across diverse configurations. For comparison, we select DPM++ SDE~\citep{lu2022dpm}, DPM++ 2M~\citep{lu2022dpm}, and EDM~\citep{karras2022elucidating}, using the prompt \textit{``a photo of a cat''}. The results, presented in Fig.~\ref{fig:supp_scheduler}, show that our refined noise consistently produces reliable outputs regardless of the denoising timestep or scheduler. This demonstrates the robustness of the \ourmodel across diverse schedulers and denoising step configurations.

\clearpage


\end{document}